\documentclass[10pt,twocolumn,letterpaper]{article}

\usepackage[pagenumbers]{cvpr} 

\usepackage{amsmath,amsfonts,bm}

\def\eqref#1{equation~\ref{#1}}

\def\1{\bm{1}}

\DeclareMathAlphabet{\mathsfit}{\encodingdefault}{\sfdefault}{m}{sl}
\SetMathAlphabet{\mathsfit}{bold}{\encodingdefault}{\sfdefault}{bx}{n}

\newcommand{\softmax}{\mathrm{softmax}}

\usepackage{url}

\usepackage{adjustbox}
\usepackage{array}
\usepackage{graphicx, color}
\usepackage{blindtext}

\usepackage{booktabs}
\usepackage{soul}

\usepackage{bm}
\usepackage{bbm} 
\usepackage{array} 
\usepackage[dvipsnames]{xcolor}
\usepackage[safe]{tipa} 
\usepackage[misc,geometry]{ifsym} 

\usepackage{comment}
\usepackage{amsmath,amssymb} 
\usepackage{color}
\usepackage{ragged2e}
\usepackage{xcolor}         
\usepackage{graphicx}

\usepackage{subcaption}
\usepackage{multirow}
\usepackage{xspace}
\usepackage{caption}
\usepackage{float}

\usepackage{url}            
\usepackage{booktabs}       
\usepackage{amsmath}
\usepackage{amsfonts}       

\usepackage{nicefrac}       
\usepackage{microtype}      
\usepackage{nicefrac}       

\newcommand{\x}{\mathbf{x}}

\newcommand{\y}{\mathbf{y}}
\newcommand{\z}{\mathbf{z}}
\newcommand{\bv}{\mathbf{v}}

\newcommand{\btheta}{\boldsymbol{\theta}}

\newcommand*\rot{\rotatebox{90 }}

\usepackage{colortbl} 
\definecolor{mColor1}{rgb}{0.9,0.9,0.9}
\definecolor{mColor2}{rgb}{0.95,0.95,0.95}
\definecolor{non-photoblue}{rgb}{0.64, 0.87, 0.93}
\definecolor{lightblue}{rgb}{0.81, 0.94, 1.0}
\usepackage{tikz}
\usepackage{pifont}
\newcommand{\ymark}{\ding{51}}

\usepackage{wrapfig}

\usepackage{etoc}
\usepackage{algorithm}
\usepackage{algorithmic}
\usepackage{xcolor}   
\definecolor{mydarkblue}{rgb}{0,0.08,0.45}

\definecolor{mColor1}{rgb}{0.9,0.9,0.9}
\definecolor{mColor2}{rgb}{0.95,0.95,0.95}
\definecolor{non-photoblue}{rgb}{0.64, 0.87, 0.93}
\definecolor{lightblue}{rgb}{0.81, 0.94, 1.0}
\definecolor{lightorange}{rgb}{0.965, 0.835, 0.71}

\makeatletter
\def\@fnsymbol#1{\ensuremath{\ifcase#1\or \textsuperscript{~\Letter}\or \ddagger\or
   \mathsection\or \mathparagraph\or \|\or **\or \dagger\dagger
   \or \ddagger\ddagger \else\@ctrerr\fi}}
\makeatother

\newcommand{\tableCellHeight}{1}
\newcommand{\tabstyle}[1]{
  \setlength{\tabcolsep}{#1}
  \renewcommand{\arraystretch}{\tableCellHeight}
  \centering
  \small
}

\definecolor{cvprblue}{rgb}{0.21,0.49,0.74}
\usepackage[pagebackref,breaklinks,colorlinks,citecolor=cvprblue]{hyperref} 

\def\paperID{8198 } 
\def\confName{CVPR}
\def\confYear{2024}

\title{Any-Shift Prompting for Generalization over Distributions}

\author{
\hspace{4mm}Zehao Xiao\textsuperscript{1} 
\hspace{8mm}Jiayi Shen\textsuperscript{1} 
\hspace{8mm}Mohammad Mahdi Derakhshani\textsuperscript{1} \\
\hspace{4mm}Shengcai Liao\textsuperscript{2} 
\hspace{8mm}Cees G. M. Snoek\textsuperscript{1} \\
\hspace{3mm}\textsuperscript{1}University of Amsterdam \hspace{3mm}\textsuperscript{2}Core42
}

\begin{document}
\maketitle

\begin{abstract}
Image-language models with prompt learning have shown remarkable advances in numerous downstream vision tasks. Nevertheless, conventional prompt learning methods overfit their training distribution and lose the generalization ability on test distributions. To improve generalization across various distribution shifts, we propose any-shift prompting: a general probabilistic inference framework that considers the relationship between training and test distributions during prompt learning. We explicitly connect training and test distributions in the latent space by constructing training and test prompts in a hierarchical architecture. Within this framework, the test prompt exploits the distribution relationships to guide the generalization of the CLIP image-language model from training to any test distribution. To effectively encode the distribution information and their relationships, we further introduce a transformer inference network with a pseudo-shift training mechanism. The network generates the tailored test prompt with both training and test information in a feedforward pass, avoiding extra training costs at test time. Extensive experiments on twenty-three datasets demonstrate the effectiveness of any-shift prompting on the generalization over various distribution shifts. 
\end{abstract}

\section{Introduction}

\begin{figure}[t]
\centering
\includegraphics[width=0.99\linewidth]{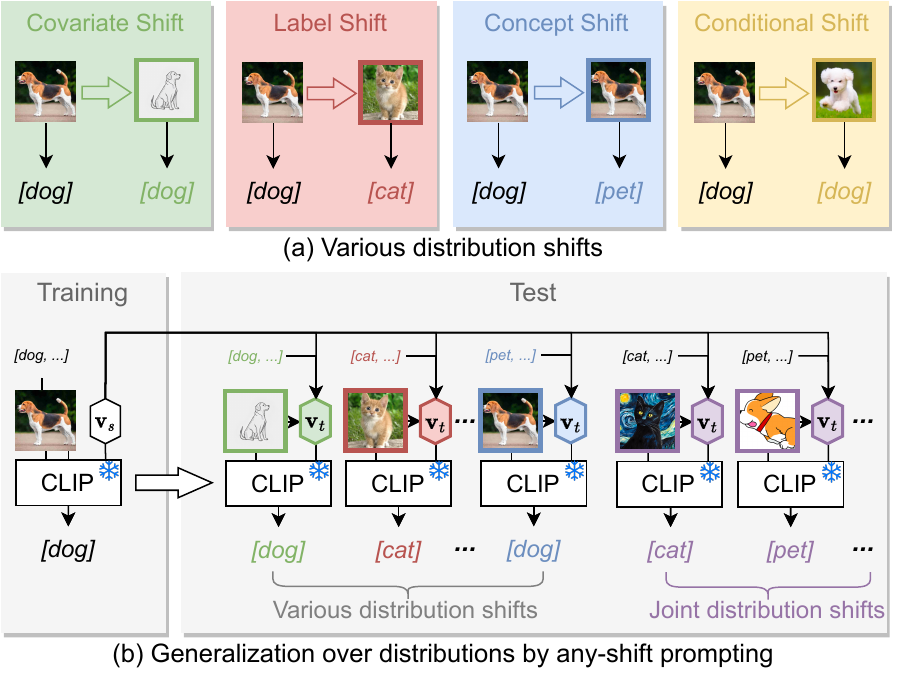}
\vspace{-3mm}
\caption{\textbf{Any-shift prompting.} (a) Various distribution shifts in real-world applications. (b) We propose any-shift prompting that aggregates training and test information for jointly handling individual distribution shifts and their combinations. 
}
\label{fig1: shifts}
\vspace{-7mm}
\end{figure}

Recent image-language foundation models like CLIP \cite{radford2021learning} show remarkable advances in various computer vision tasks. Benefiting from large image-text pairing datasets for pre-training, these models perform well when adapting to downstream tasks by manual prompts \cite{liu2023pre, novack2023chils, roth2023waffling, ramesh2022hierarchical} and prompt learning \cite{zhou2022learning, zhou2022conditional}. However, it is difficult for conventional prompt learning approaches to handle distribution shifts in downstream tasks \cite{shu2022test, derakhshani2023bayesian}. The learned prompts usually overfit their training data, leading to performance degradation on unseen test distributions. 

To improve generalization of prompt learning, recent methods introduce uncertainty into the learnable prompt \cite{derakhshani2023bayesian} or fine-tune the prompt on each test sample with extra unsupervised optimizations \cite{shu2022test, samadh2023align}. Nevertheless, these methods do not explicitly consider the relationships between training and test distributions of the downstream tasks. However, in real-world applications, the distribution shifts are usually complex and unpredictable, where models may encounter different distribution shifts (Figure~\ref{fig1: shifts} (a)), and even their combinations. Hence, we deem it crucial to explore the relationships between training and test distributions for the generalization of prompting across different distribution shifts. To this end, we make three contributions in this paper. 

First, we propose any-shift prompting, a general probabilistic inference framework that can explore distribution relationships in prompt learning. Specifically, we introduce probabilistic training and test prompts in a hierarchical architecture to explicitly connect the training and test distributions. Within this framework, the test prompt encodes the test information and the relationships of the training and test distributions, thereby improving the generalization ability on various test distributions (Figure~\ref{fig1: shifts} (b)).

Second, we propose a pseudo-shift training mechanism, where the hierarchical probabilistic model learns the ability to encode distribution relationships by simulating distribution shifts. Consequently, at test time, our method generalizes to any specific distribution by generating a tailored prompt on the fly in just one feedforward process, without the need for re-learning or fine-tuning. 

Third, to effectively and comprehensively encode the distribution information and their relationships, we design a transformer inference network for prompt generation. The transformer takes test information of both image and label space features, as well as the training prompts, as inputs. It then aggregates the training and test information and their relationships into the test-specific prompt. The test prompt is utilized to guide both the feature extraction and classification processes to generate test-specific features and classifiers, which bolsters robust predictions across distribution shifts.

We validate our method through extensive experiments on twenty-three benchmarks with various distribution shifts, including covariate shift, label shift, conditional shift, concept shift, and even joint shift. The results demonstrate the effectiveness of the proposed method on generalization across various distribution shifts.

\section{Preliminary}

We propose any-shift prompting based on CLIP \cite{radford2021learning} to handle various distribution shifts in a general way. Here we provide the technical background on CLIP as well as definitions of distribution shifts considered.

\vspace{2mm}
\noindent
\textbf{CLIP model.} 
Contrastive Language-Image Pre-training (CLIP) \cite{radford2021learning} consists of an image encoder $f_{\Phi_{I}}(\x)$ and a text encoder $f_{\Phi_{T}}(\mathbf{l)}$, which are trained by a contrastive loss on a large dataset of image-language ($\x, \mathbf{l}$) pairs. For a downstream classification task with an input image $\x$ and a set of class names $\mathcal{Y} {=} \{ c_{i} \}_{i {=} 1}^C$, the image feature is extracted by $\z {=} f_{\Phi_{I}}(\x)$ and the classifiers are composed of a set of text features $\{ \mathbf{t}_i \}_{i {=} 1}^C$, where $\mathbf{t}_i {=} f_{\Phi_{T}}(\mathbf{l}_i)$. Here, $\mathbf{l}_i$ is a manually crafted prompt to describe the corresponding class name $c_{i}$, e.g., ``\textit{an image of a [class]}.''
Thus, the prediction function of the CLIP model for downstream tasks without fine-tuning is formulated as: 
\begin{equation}
    p(\y|\x, \mathcal{Y}) {=} \softmax(\z^{\top} \mathbf{t}).
\label{eq: clip}
\end{equation} 
This enables the pre-trained CLIP model to handle zero-shot learning classification in various downstream tasks.

\vspace{2mm}
\noindent
\textbf{Distribution shifts.}
A data distribution is generally denoted as $p(\x, \y)$, which is a joint distribution of the input data $\x$ and the label $\y$. The models are usually trained on a training distribution $p(\x_s, \y_s)$ and then deployed on test distributions $p(\x_t, \y_t)$.
In real-world applications, differences between the training and test distributions are known as the joint distribution shift:
\begin{equation}
\label{eq: ds}
p(\x_s, \y_s) \neq p(\x_t, \y_t).
\end{equation}

\begin{table}[t]
\centering
	\resizebox{1\columnwidth}{!}{%
\begin{tabular}{lc}
\toprule
Joint distribution shift &  {$p(\x_s, \y_s) \neq p(\x_t, \y_t)$} \\ \midrule
\rowcolor{mColor1}
\multicolumn{2}{c}{Partial distribution shifts} \\
Covariate shift & $p(\x_s) \neq p(\x_t)$ ~~ $p(\y_s|\x_s) = p(\y_t|\x_t)$\\
Label shift & $p(\y_s) \neq p(\y_t)$ ~~ $p(\x_s|\y_s) = p(\x_t|\y_t)$\\
{Concept shift} & $p(\x_s) = p(\x_t)$ ~~ $p(\y_s|\x_s) \neq p(\y_t|\x_t)$\\
{Conditional shift} & $p(\y_s) = p(\y_t)$ ~~ $p(\x_s|\y_s) \neq p(\x_t|\y_t)$ \\ 
\bottomrule
\end{tabular}
}
\vspace{-2mm}
\caption{
\textbf{Common distribution shifts.} 
The joint distribution shift is usually decomposed into four partial shifts, which are investigated individually in the literature. By contrast, we focus in this paper on various shifts and even consider their combinations.
}\label{settings}
\vspace{-6mm}
\end{table}

\noindent
\textbf{Common distribution shifts in the literature.}
Due to the joint distribution shift, the performance of the trained model degrades on the test data \cite{wang2021tent, lim2023ttn}, sometimes significantly so. 
Since the joint distribution shift is complex,  previous methods limit the scope of the problem and simplify the joint distribution shift to different partial distribution shifts.
From a Bayesian perspective, the joint distribution is decomposed into $p(\x, \y) {=} p(\x)p(\y|\x) {=} p(\y)p(\x|\y)$. 
According to the different components in the decomposition, we summarize the partial distribution shifts into four different definitions in Table~\ref{settings} and detail them one by one. 

\textit{Covariate shift} \cite{sun2020test, li2017deeper, schneider2020improving} assumes the distribution shifts occur only in the input space $p(\x)$ while the labels given the input features $p(\y|\x)$ remain the same, e.g., by image corruptions \cite{hendrycks2019benchmarking} or changing image styles \cite{li2017deeper, peng2019moment}.
Covariate shift is widely investigated by domain generalization \cite{zhou2022domain, li2017deeper, xiao2021bit} and domain adaptation methods \cite{wang2021tent, lim2023ttn}. \textit{Label shift} focuses on the opposite problem, where the label distributions $p(\y)$ are different, but the label-conditional distributions $p(\x|\y)$ are the same \cite{tachet2020domain, roberts2022unsupervised}. 
Previous methods generate datasets with uniform distribution $p(\y)$ during training and different distributions at test time \cite{guo2020ltf, azizzadenesheli2019regularized, wu2021online}.  
The classification of unknown classes can be treated as a specific and worse case of the label shift \cite{liu2021adversarial, shen2022association,  zhou2022conditional}, where $p(\y) {=} 0$ for the unknown classes.
\textit{Concept shift} treats the distribution of input $p(\x)$ the same while the conditional distributions $p(\y|\x)$ are different, indicating different annotation methods for the same data distribution \cite{liu2022deep}.
\textit{Conditional shift} assumes the label distribution is the same while the conditional distribution $p(\x|\y)$ are different \cite{liu2021adversarial, zhang2013domain, gong2016domain}, where different classes can have their own shift protocols on the input data, e.g., sub-population problems \cite{santurkar2020breeds, lee2022surgical}. 

\vspace{2mm}
\noindent
\textbf{Distribution shifts in this paper.}
Conventional prompting methods \cite{zhou2022learning, zhou2022conditional} learn the prompt on the training distribution of the downstream task, which is easy to overfit and vulnerable to the above shifts \cite{derakhshani2023bayesian, shu2022test}. 
Moreover, in real-world scenarios, all distribution shifts may happen unpredictably, and even simultaneously. Hence, we propose to encode test information and the training-test relationships for generalization over distributions. Our method is not designed for specific partial distribution shifts. Instead, it is proposed to handle various shifts, even when they occur simultaneously.

\section{Any-Shift Prompting}

\subsection{Prompt modeling}
\begin{figure}[t]
\centering
\includegraphics[width=0.99\linewidth]{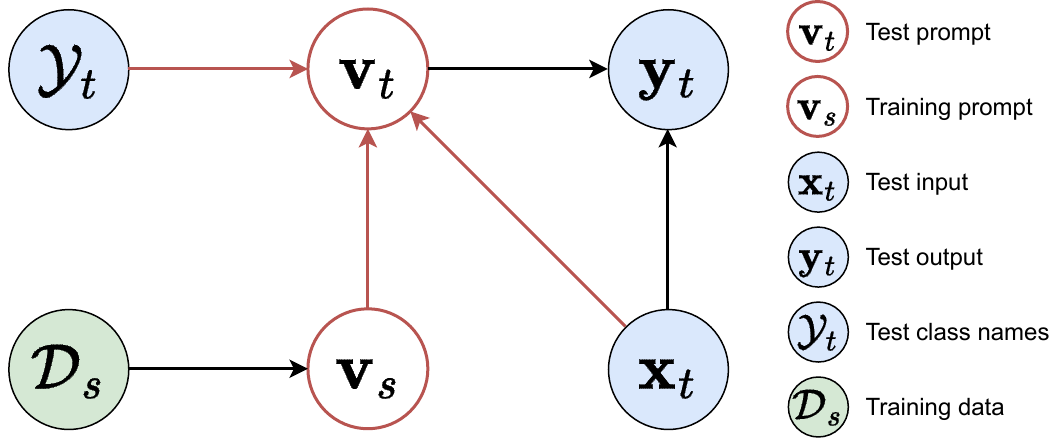}
\vspace{-2mm}
\caption{\textbf{Graphical model for any-shift prompting.} We introduce probabilistic training and test prompts in a hierarchical inference framework to explore distribution relationships.}
\label{fig: graphmodel}
\vspace{-4mm}
\end{figure}

We propose any-shift prompting, a general probabilistic inference framework to explore distribution relationships.
Specifically, we introduce training and test prompts as latent variables in a hierarchical architecture.
The graphical model of our method is provided in Figure~\ref{fig: graphmodel}.

\vspace{1mm}
\noindent
\textbf{Training prompt.}
The intuitive idea of adapting the CLIP model is to inject the downstream training data $\mathcal{D}_s$ in a training prompt for prediction (eq.~\ref{eq: clip}). 
$\mathcal{D}_s$ consists of training input-output pairs sampled from the distribution $p(\x_s, \y_s)$. 
The predictive function of CLIP for the test distribution $p(\x_t, \y_t)$ is then formulated as:
\begin{equation}
p_{\Phi}(\y_t|\x_t, \mathcal{Y}_t, \mathcal{D}_s) \propto p_{\Phi}(\y_t|\x_t, \bv_s, \mathcal{Y}_t)p(\bv_s|\mathcal{D}_s),
\end{equation} where $\Phi$ denotes the frozen parameters of the image and text encoders of the CLIP model. Here $\bv_s$ is the training prompt that encodes the training downstream task information, which improves the performance of the CLIP model on the training distribution. However, the prompt $\bv_s$ usually overfits the training data, which may not benefit and even harm the prediction on the unseen test distribution due to the distribution shifts at test time.  

\vspace{-2mm}
\paragraph{Probabilistic test prompt.}
To generalize across distribution shifts in downstream tasks at test time, we further introduce a probabilistic test prompt within a hierarchical Bayes framework to encode the information of test distributions.
Specifically, the test prompt $\bv_t$ is inferred from the training prompt $\bv_s$ and the accessible test information, i.e., a test image $\x_t$ and the class names $\mathcal{Y}_t$. 
To build the connections between the training and test prompts, we take the training prompt $\bv_s$ as a condition for the generation of the test prompt.
This enables the method to generate the test prompt across different shifts by considering the relationships between training and test distributions and exploring relevant training information.
By introducing $\bv_t$, the CLIP prediction function is formulated as:
\begin{equation}
\small
\label{eq: eq2_prior}
\begin{aligned}
    & p_{\Phi, \theta}(\y_t|\x_t, \mathcal{Y}_t, \mathcal{D}_s) \\
    & = \int \int p_{\Phi}(\y_t|\x_t, \bv_t, \mathcal{Y}_t) p_{\btheta}(\bv_t|\bv_s, \x_t, \mathcal{Y}_t) p(\bv_s|\mathcal{D}_s) d\bv_t d\bv_s,
\end{aligned}
\end{equation}
where $\btheta$ denotes the learnable inference network for the test prompt. With the probabilistic test prompt, we provide a general way to incorporate the training and test information, as well as their relationships, into the prediction of the CLIP model, enabling it to generalize on any test distribution.

\vspace{1mm}
\noindent
\textbf{Variational test prompt.}
To optimize the model for generating the probabilistic test prompt in eq.~(\ref{eq: eq2_prior}), we use variational inference to approximate the true posterior $p(\bv_t, \bv_s|\mathcal{D}_t, \mathcal{Y}_t, \mathcal{D}_s)$, which is factorized as:
\begin{equation}
q_{\btheta}(\bv_t, \bv_s|\mathcal{D}_t,\mathcal{Y}_t, \mathcal{D}_s) {=} q_{\btheta}(\bv_t|\bv_s, \mathcal{D}_t, \mathcal{Y}_t) p(\bv_s|\mathcal{D}_s),
\label{eq: varational}
\end{equation}
where $\mathcal{D}_t$ consists of test input-output pairs sampled from the test distribution $p(\x_t, \y_t)$. The variational posterior of the test prompt shares the same inference model $\btheta$ with its prior.
By integrating eq.~(\ref{eq: varational}) into eq.~(\ref{eq: eq2_prior}), we derive the evidence lower bound (ELBO) of the predictive function as: 
\begin{equation}
\small
\label{eq: eq3_elbo}
\begin{aligned}
    & \log p_{\Phi, \btheta}(\y_{t}|\x_{t}, \mathcal{Y}_{t}, \mathcal{D}_s) 
    \geq \mathbb{E}_{q_{\btheta}(\bv_{t}, \bv_s)} \big [ \log p_{\Phi}(\y_t | \x_t, \bv_{t}, \mathcal{Y}_{t}) \big ] \\
    & ~~~~~ - \mathbb{D}_\mathrm{KL} \big [ q_{\btheta}(\bv_{t}|\bv_s, \mathcal{D}_t, \mathcal{Y}_{t}) || p_{\btheta}(\bv_{t}|\bv_s, \x_{t}, \mathcal{Y}_{t})  \big ].
\end{aligned}
\end{equation}
The variational posterior of the test prompt $q_{\btheta}(\bv_{t})$ encodes more input-output information of the test distribution and their relationships, yielding a more representative test prompt for better generalization on the test distributions. 
We provide the step-by-step derivations in the supplemental material.

Notably, the variational posteriors and ELBO are intractable since large numbers of test samples and their ground truth labels in $\mathcal{D}_t$ are usually unavailable at test time. 
Thus, in the next section, we propose a pseudo-shift training setup to approximate the ELBO for any-shift prompting.

\begin{figure*}[t]
\centering
\includegraphics[width=0.99\linewidth]{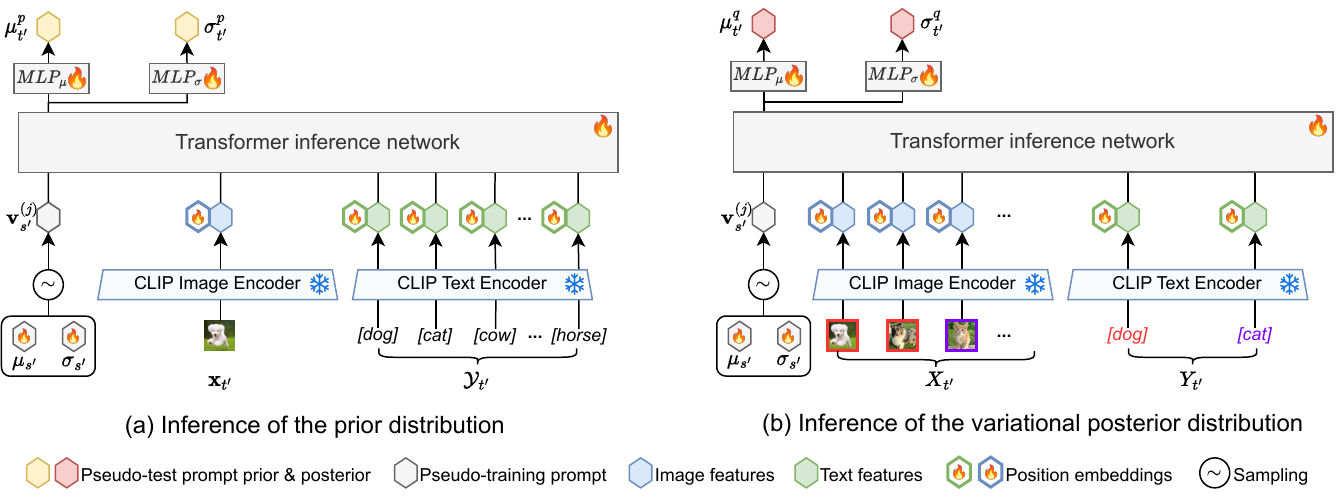}
\vspace{-4mm}
\caption{
\textbf{Transformer inference network of the pseudo-test prompt.} 
The prior (a) of the pseudo-test prompt is inferred by aggregating the pseudo-training prompt, a single image, and all class names of the pseudo-test distribution. The posterior (b) is inferred from the shared pseudo-training prompt, a batch of pseudo-test images, and corresponding class names.
Therefore, the posterior incorporates more pseudo-test information and relationships and guides the prior to learn the same knowledge by KL divergence.
The image and text encoders of CLIP are frozen. Only the shared transformer, pseudo-training prompt distribution, and MLP networks are trainable, saving training costs. 
}
\label{fig: transformer}
\vspace{-5mm}
\end{figure*}

\subsection{Training and inference}
\label{training}

\paragraph{Pseudo-shift training mechanism.}
To approximate the intractable ELBO in eq.~(\ref{eq: eq3_elbo}), we develop a pseudo-shift training mechanism. Specifically, the mini-batch data in the current iteration is treated as the pseudo-test data $\mathcal{D}_{t'}$ from the pseudo-test distribution $p(\x_{t'}, \y_{t'})$. Likewise, the mini-batches in previous iterations are treated as the pseudo-training data $\mathcal{D}_{s'}$ from the pseudo-training distribution $p(\x_{s'}, \y_{s'})$. In this case, the ground truth labels of the pseudo-test data are available during training. We then approximate the ELBO and obtain the optimization function for any-shift prompting as:
\begin{equation}
\label{eq: eq4_loss}
\begin{aligned}
    & \mathcal{L} {=} - \mathbb{E}_{q_{\btheta}(\bv_{t'}, \bv_{s'})} \big [ \log p_{\Phi}(\y_{t'}|\x_{t'}, \bv_{t'}, \mathcal{Y}_{t'}) \big ] \\
    & + \mathbb{D}_\mathrm{KL} \big [ q_{\btheta}(\bv_{t'}|\bv_{s'}, \mathcal{D}_{t'}, \mathcal{Y}_{t'}) || p_{\btheta}(\bv_{t'}|\bv_{s'}, \x_{t'}, \mathcal{Y}_{t'}) \big ],
\end{aligned}
\end{equation} 
where $\bv_{t'}$ and $\bv_{s'}$ denote the pseudo-test and pseudo-training prompts, respectively. In practice, we assume the prompts follow the standard Gaussian distributions. The negative log-likelihood in eq.~(\ref{eq: eq4_loss}) is implemented by a cross-entropy loss. 
The mini-batch training mechanism mimics the distribution shifts and trains the any-shift prompting to handle the distribution shifts during training, where the model never accesses any test data.
Minimizing the KL terms encourages the prior to implicitly learn more comprehensive pseudo-test information from the variational posterior, which aggregates more data information together with the ground truth labels.

\vspace{1mm}
\noindent
\textbf{Transformer inference network.}
The pseudo-test prompt in eq.~(\ref{eq: eq4_loss}) is inferred from: the pseudo-training information in $\bv_{s'}$, the pseudo-test image $\x_{t'}$, and the class names $\mathcal{Y}_{t'}$.
To better aggregate the different information sources and consider their relationships, we introduce a transformer inference network to generate the pseudo-test prompt.

In our model, the prior $p_{\btheta}(\bv_{t'}|\bv_{s'}, \x_{t'}, \mathcal{Y}_{t'})$ and variational posterior $q_{\btheta}(\bv_{t'}|\bv_{s'}, \mathcal{D}_{t'}, \mathcal{Y}_{t'})$ of the pseudo-test prompt share the same inference network to encode the different conditions. Compared with the prior, the variational posterior has access to one batch of pseudo-test images with the corresponding ground-truth labels. Figure~\ref{fig: transformer} illustrates the deployment of the shared transformer inference network. In the following, we provide the detailed inference of the prior and variational posterior. 

As shown in Figure~\ref{fig: transformer} (a), the prior of the pseudo-test prompt is generated by the pseudo-training prompt $\bv_{s'}$, the pseudo-test image $\x_t'$, and class names $\mathcal{Y}_t'$. Specifically, we sample a pseudo-training prompt $\bv_{s'}^{(j)}$ from a Gaussian distribution $\mathcal{N}(\bv_{s'}; \mu_{s'}, \sigma_{s'})$ by the reparameterization trick~\cite{kingma2013auto}. The mean $\mu_{s'}$ and variance $\sigma_{s'}$ are two sets of parameters trained with the pseudo-training data $\mathcal{D}_{s'}$ in the previous iterations. The pseudo-test image is fed into the fixed CLIP image encoder to get the image feature $f_{\Phi_I}(\x_{t'})$. The class names of the pseudo-test distribution are processed by the fixed text encoder to extract the textual features $f_{\Phi_T}(\mathcal{Y}_{t'})$.  After the pre-processing, we take the sampled pseudo-training prompt, pseudo-test image feature, and textual features as input tokens of our transformer inference network to generate the prior of the pseudo-test prompt:
\begin{equation}
\label{eq: transp}
[\widetilde{\bv}_{t'}^p; \cdot ; \cdot] = \texttt{Trans} ([\bv^{(j)}_{s'}; f_{\Phi_I}(\x_{t'}); f_{\Phi_T}(\mathcal{Y}_{t'})]), 
\end{equation}
\begin{equation}
\label{eq: mlp}
\mu_{t'}^p = \texttt{MLP}_{\mu} (\widetilde{\bv}_{t'}^p), ~~~~ \sigma_{t'}^p = \texttt{MLP}_{\sigma} (\widetilde{\bv}_{t'}^p),
\end{equation}
\begin{equation}
\label{eq: disp}
p_{\btheta}(\bv_{t'}  | \bv_{s'}, \x_{t'}; \mathcal{Y}_{t'})  =  \mathcal{N}(\bv_{t'}; \mu_{t'}^p, \sigma_{t'}^p).
\end{equation}
The prior of the pseudo-test prompt follows the Gaussian distribution in eq.~(\ref{eq: disp}), whose mean and variance are obtained by two MLP networks on the output of the transformer $\widetilde{\bv}_{t'}^p$.

In Figure~\ref{fig: transformer} (b), with the pseudo-test data $\mathcal{D}_{t'}$, the variational posterior learns more distribution information as well as the relations between inputs and outputs. 
To be clearer, we rewrite the variational posterior $q_{\btheta}(\bv_{t'}|\bv_{s'}, \mathcal{D}_{t'}, \mathcal{Y}_{t'})$ as $q_{\btheta}(\bv_{t'}|\bv_{s'}, X_{t'}, Y_{t'})$,
where $X_{t'}$ contains a batch of pseudo-test images in $\mathcal{D}_{t'}$ and $Y_{t'}$ consists of the ground truth class names of $X_{t'}$ in $\mathcal{Y}_{t'}$.
Hence, the shared transformer takes all image features and their corresponding label features as input tokens to infer the variational posterior:
\begin{equation}
\label{eq: transq}
[\widetilde{\bv}_{t'}^q; \cdot ; \cdot] = \texttt{Trans} ([\bv^{(j)}_{s'}; f_{\Phi_I}(X_{t'}); f_{\Phi_T}(Y_{t'})]), 
\end{equation}
\begin{equation}
\label{eq: mlpq}
\mu_{t'}^q = \texttt{MLP}_{\mu} (\widetilde{\bv}_{t'}^q), ~~~~ \sigma_{t'}^q = \texttt{MLP}_{\sigma} (\widetilde{\bv}_{t'}^q),
\end{equation}
\begin{equation}
\label{eq: disq}
q_{\btheta}(\bv_{t'} | \bv_{s'}, \mathcal{D}_{t'}, \mathcal{Y}_{t'})  =  \mathcal{N}(\bv_{t'}; \mu_{t'}^q, \sigma_{t'}^q).
\end{equation}
With the inferred pseudo-test prompt, we take its samples from the variational posterior as the input tokens for both image and text encoders of CLIP to make predictions during training. 
Thus, although the encoders are fixed, the image and textual features are generalized by utilizing the distribution information in the prompts during the feature extraction and classification procedure, enabling the method to handle different distribution shifts.

\vspace{1mm}
\noindent
\textbf{Prediction.}
At test time, we make predictions on each test image $\x_t$ with the test prompt generated by the transformer inference network. Since the test data and labels in $\mathcal{D}_{t}$ are unavailable, the variational posterior becomes intractable. Thus, we sample the test prompt  $\bv_t^{(i)}$ from the prior distribution $p_{\btheta}(\bv_t|\bv_s^{(j)}, \x_t, \mathcal{Y}_t)$, where $\bv_s^{(j)}$ is a sample of the training prompt following $p(\bv_s|\mathcal{D}_s)$. 
$\bv_t^{(i)}$ is then introduced into both the image and text encoders of the CLIP model for generalization and prediction as:
\begin{equation}
\small
\label{eq: test}
\begin{aligned}
    p_{\Phi}(\y_t|\x_t, \mathcal{Y}_t, \mathcal{D}_s) = \frac{1}{N_t} \frac{1}{N_s} \sum_{i=1}^{N_t}  & \sum_{j=1}^{N_s} p_{\Phi}(\y_t|\x_t, \bv_t^{(i)}, \mathcal{Y}_t), \\
    \bv_t^{(i)} \sim p_{\btheta}(\bv_t|\bv^{(j)}_s, \x_t, \mathcal{Y}_t), ~~~ & ~~~
    \bv_s^{(j)} \sim  p(\bv_s|\mathcal{D}_s). \\
\end{aligned}
\end{equation}
Although the test data and their labels are not available at test time, the information in each test sample and all class names in the vocabulary of the test task are available to infer the prior of the test prompt.
The ability to encode test information from a single test image and the class vocabulary is learned during training by minimizing the KL divergence between the prior and posterior. Note the CLIP image encoder and text encoder are always frozen. 
Only the test prompt changes for different test distributions by aggregating the training and test information in each test sample $\x_t$ and the class names $\mathcal{Y}_t$. In this case, we utilize the original generalization ability of CLIP to generate the test prompt for generalization on downstream tasks across various distribution shifts.

\section{Related Work}

\vspace{1mm}
\noindent
\textbf{Prompt learning.} 
Image-language foundation models such as CLIP \cite{radford2021learning} and ALIGN \cite{jia2021scaling} achieve significant advances in various downstream tasks. To adapt the foundation models to downstream tasks, adapter \cite{gao2023clip} and prompt learning methods \cite{lester2021power, li2021prefix, zhou2022learning} are proposed.
Zhou \etal \cite{zhou2022learning} propose a learnable prompt as the input of the language model in CLIP. To avoid forgetting the original knowledge in the CLIP model, Zhu \etal \cite{zhu2023prompt} and Yao \etal \cite{yao2023visual} guide prompt learning with hand-crafted prompts. Instead of generating prompts for the language model, Bahng \etal \cite{bahng2022exploring} introduce prompting of the image model.
Khattak \etal \cite{khattak2023maple} learn a joint prompt for both image and language encoders. Zhou \etal \cite{zhou2022conditional} introduce the imaging conditions into the language prompt to enhance the generalization ability of zero-shot performance. To further improve the generalization ability, Derakhshani \etal \cite{derakhshani2023bayesian}
propose Bayesian prompt learning, which considers the uncertainty in the learned prompts for zero-shot generalization.
Shu \etal \cite{shu2022test} and Hassan \etal \cite{samadh2023align} fine-tune the prompt at test time to a specific distribution. 
We also improve the generalization of prompt learning. 
Different from previous methods that consider uncertainty or fine-tune the prompt for specific distributions, we propose any-shift prompting that explicitly explores distribution information and relationships within a hierarchical probabilistic framework. The method generates the test-specific prompt on the fly for any test distribution.

\vspace{1mm}
\noindent
\textbf{Distribution shift generalization.}
Domain generalization \cite{muandet2013domain, li2018domain, zhou2022domain, gulrajani2020search} and domain adaptation \cite{long2015learning, wang2018deep, liang2020we, yi2023source} are the most widely investigated methods for handling distribution shifts. Some domain generalization methods train invariant models on the training distributions \cite{arjovsky2019invariant, xiao2021bit, motiian2017unified}, which are assumed to be invariant on the test distributions also. To further improve the generalization ability, some methods \cite{li2018metalearning, dou2019domain, balaji2018metareg} introduce meta-learning in domain generalization to mimic domain shifts during training. In this paper, we also simulate the distribution shift by a pseudo-shift training mechanism, which uses different mini-batches as distributions. To better utilize the test information for generalization without accessing the test data during training, Sun \etal \cite{sun2020test} and Wang \etal \cite{wang2021tent} propose test-time adaptation, which fine-tunes the trained model on test data with self-supervised losses. The method is followed by many methods \cite{zhang2021memo, niu2022efficient, goyal2022test, niu2023towards, liu2021ttt++} due to its good generalization ability on covariate shift. 
In addition, test-time adaptation is also investigated with other methods like normalization statistics re-estimation \cite{schneider2020improving, lim2023ttn}, or classifier adjustment \cite{iwasawa2021test, xiao2022learning, zhang2023adanpc}.
Most of these methods focus on covariate shift \cite{wang2021tent, xiao2022learning, goyal2022test, dubey2021adaptive}, such as changes of the image styles \cite{li2017deeper, peng2019moment} and corruptions \cite{hendrycks2019benchmarking}.
Some other methods work on the conditional shift \cite{liu2021adversarial, zhang2013domain, gong2016domain, garg2023rlsbench} or label shift \cite{zhang2013domain, tachet2020domain, park2023label, garg2023rlsbench}.
We also utilize the test information for generalization, but without any test-time optimization. Different from the previous methods, we explicitly bridge the training and test information and explore their relationships to address various distribution shifts in a general way.

\section{Experiments}

\begin{table*}[t]
\begin{center}
\vspace{-2mm}
	\resizebox{1.98\columnwidth}{!}{%
\begin{tabular}{l llll  llll}
\toprule
 Method   & {PACS} & {VLCS} & {Office-Home} & {DomainNet} & {ImageNet-V2} & {ImageNet-S} & {ImageNet-A} & {ImageNet-R}\\ \midrule
 \rowcolor{mColor1}
 \multicolumn{9}{l}{\textbf{Prompting without test-time optimization}} \\
CLIP \cite{radford2021learning} & 96.13 & 81.43 & 80.35  & 
 54.08 & 60.83 & 46.15 & 47.77 & 73.96 \\
CLIP-D \cite{radford2021learning} & 96.65 & 80.70 & 81.51  & 
 56.24 & - & - & - & - \\
CoOp \cite{zhou2022learning} & 96.45 & 82.51 & 82.12 & 58.82 & {64.20} & 47.99 & 49.71 & 75.21 \\
CoCoOp \cite{zhou2022conditional} & 97.00 & 83.89 & 82.77  & 59.43 & 64.07 & {48.75} & {50.63} & {76.18}  \\
DPL \cite{zhang2021domain}  & 97.07 & 83.99 & 83.00 & 59.86 & - & - & - & -  \\
BPL~\cite{derakhshani2023bayesian} & - & - & - & - & 64.23 & 49.20 & \textbf{51.33} & 77.00 \\
\rowcolor{lightblue}
\textit{\textbf{This paper}}  & \textbf{98.16} \scriptsize{$\pm$ 0.4} & \textbf{86.54} \scriptsize{$\pm$ 0.4} & \textbf{85.16} \scriptsize{$\pm$ 0.6} & \textbf{60.93} \scriptsize{$\pm$ 0.6}  & \textbf{64.53} \scriptsize{$\pm$ 0.2} & \textbf{49.80} \scriptsize{$\pm$ 0.5} & \textbf{51.52} \scriptsize{$\pm$ 0.6} & \textbf{77.56} \scriptsize{$\pm$ 0.4} \\ 
\midrule
\rowcolor{mColor1}
\multicolumn{9}{l}{\textbf{Prompting with test-time optimization}} \\
TPT \cite{shu2022test}  & 97.25 & 84.33 & 83.45 & 59.90 & 63.45 & 47.94 & {54.77} & 77.06 \\
CoOp + TPT \cite{shu2022test} & {97.85} & 85.06 & 84.32 & {60.65} & \textbf{66.83} & 49.29 & 57.95 & 77.27  \\
CoCoOp + TPT \cite{shu2022test} & {97.95} & 85.55 & 84.54  & {60.44} & 64.85 & 48.47 & \textbf{58.47}  & 78.65 \\ 
\rowcolor{lightblue}
\textit{\textbf{This paper + TPT}}   & \textbf{98.47} \scriptsize{$\pm$ 0.4}  &  \textbf{86.98} \scriptsize{$\pm$ 0.4} & \textbf{86.00} \scriptsize{$\pm$ 0.8}  & \textbf{61.75} \scriptsize{$\pm$ 0.8} & \textbf{67.08} \scriptsize{$\pm$ 0.6}   &  \textbf{50.83} \scriptsize{$\pm$ 0.6} &  \textbf{58.05} \scriptsize{$\pm$ 0.5} &  \textbf{79.23} \scriptsize{$\pm$ 0.5} \\ 
\bottomrule
\end{tabular}
}
\end{center}
\vspace{-6mm}
\caption{\textbf{Covariate shift comparison.} The experiments are conducted on eight domain generalization datasets, with average classification accuracy reported. Any-shift prompting achieves the best results compared with the original CLIP and other prompt learning methods, which demonstrates the generalization ability of our method on covariate shift. When combined with TPT's test-time optimization, promting methods in general, as well as our method improves further.
}
\label{tab:dg}
\vspace{-6mm}
\end{table*}

\paragraph{Twenty-three datasets.} To demonstrate the generalization ability of any-shift prompting, we evaluate the method on datasets with different distribution shifts. For covariate shift, we conduct experiments on the common domain generalization datasets, \texttt{PACS} \cite{li2017deeper}, \texttt{Office-Home} \cite{venkateswara2017deep}, \texttt{VLCS} \cite{fang2013video}, and \texttt{DomainNet} \cite{peng2019moment}, which contain images from different domains such as image styles.
We also evaluate the model on covariate shifts of {ImageNet} \cite{deng2009imagenet} following Zhou \etal \cite{zhou2022conditional}, where the model is trained on {ImageNet} with 16-shot images and evaluated on other variants \texttt{ImageNet-V2} \cite{recht2019imagenet}, \texttt{ImageNet-(S)ketch} \cite{wang2019learning}, \texttt{ImageNet-A} \cite{hendrycks2021natural}, and \texttt{ImageNet-R} \cite{hendrycks2021many}.
For label shift, we follow the base-to-new class generalization from Zhou \etal \cite{zhou2022learning}, with 11 datasets that cover various tasks, \texttt{ImageNet} \cite{deng2009imagenet}, \texttt{Caltech101} \cite{fei2004learning}, \texttt{OxfordPets} \cite{parkhi2012cats}, \texttt{StanfordCars} \cite{krause20133d}, \texttt{Flowers102} \cite{nilsback2008automated}, \texttt{Food101} \cite{bossard2014food}, \texttt{FGVCAircraft} \cite{maji2013fine}, \texttt{SUN397} \cite{xiao2010sun}, \texttt{DTD} \cite{cimpoi2014describing},
\texttt{EuroSAT} \cite{helber2019eurosat}, and \texttt{UCF101} \cite{soomro2012ucf101}.
For concept shift, we build a \texttt{ImageNet-Superclass} dataset, where we evaluate the ImageNet-trained model on super-classes in \cite{santurkar2020breeds}.
For conditional shift, we evaluate on the sub-population datasets \texttt{Living-17} and \texttt{Entity-30} \cite{santurkar2020breeds}, where the training and test distributions consist of the same classes with different subpopulations. 
To evaluate our method on the combination of different distribution shifts, we follow the open-domain generalization setting \cite{shu2021open} on the Office-Home dataset, which contains four domains, Art, Clipart, Product, and Real-world. We refer to it as \texttt{Open-Office-Home}, which combines covariate shift and label shift. The detailed settings are provided in the supplemental materials.

\vspace{1mm}
\noindent
\textbf{Implementation details.}
Our model consists of the pretrained image and language encoders of CLIP \cite{radford2021learning}, and the proposed transformer inference network to generate the test prompt. 
We use the ViT-B/16 \cite{dosovitskiy2020image} as the image encoder following \cite{zhou2022conditional, derakhshani2023bayesian}.
The pretrained image and language encoders of CLIP are frozen during training and inference.
To generate the prior and variational posterior of the prompt, we use a 2-layer transformer in the inference network.
As shown in Figure~\ref{fig: transformer}, the inputs of the transformer include the training prompt, the image features, and the class-name features. 
The distribution of the training prompts consists of two trainable vectors as the mean and variance respectively.
The class-name tokens are generated by the hand-crafted tokens \textit{``an image of a [class]''}.
The transformer also contains two kinds of trainable position embeddings to indicate the image and language tokens.
The introduced prompts are sampled from the corresponding distributions by the reparameterization trick~\cite{kingma2013auto}.
More detailed implementations and hyperparameters are provided in the supplemental materials. 

\subsection{Results on various distribution shifts}

\noindent
\textbf{Covariate shift.}
We conduct experiments on eight domain generalization datasets with covariate shift. The averaged results of classification accuracy for each dataset are provided in Table~\ref{tab:dg}.
We follow the leave-one-out protocol \cite{li2017deeper} for evaluation on the first four datasets, where the model evaluated on each test domain is trained on the other domains.
The detailed results on each test domain are provided in the supplemental materials.
For the last four datasets, we evaluate the same ImageNet-pretrained model on them individually. Our method outperforms the other prompt learning methods CoOp, CoCoOp, and DPL on all eight datasets.
Note that the comparisons with the other prompt learning methods are fair since we generate the test prompt and make predictions in a single feedforward pass, without any optimization or backpropagation at test time.
The proposed method also performs better on seven of the eight datasets compared with the test-time tuning method TPT, securing the second position on \texttt{ImageNet-A}.
Moreover, since the proposed method learns the prompt and transformer network only during training, it can also be combined with test-time optimization.
Then we obtain even better results, which are also competitive on \texttt{ImageNet-A},
%
indicating the effectiveness of any-shift prompting on covariate shift.

\begin{table*}[t]
    \tabstyle{4pt}
    \vspace{-2mm}
    \begin{subtable}[t]{.24\textwidth}
    \centering
    \caption{\textbf{Average over 11 datasets}.}
     \scalebox{.85}{
    \begin{tabular}{lcc|c}
    \toprule
    & Base & New & H \\
    \midrule
    CLIP  & 69.34 & {74.22} & 71.70 \\
    CoOp & \textbf{82.69} & 63.22 & 71.66 \\
    CoCoOp & 80.47 & 71.69 & {75.83} \\
    BPL  & 80.10 & {74.94} & {77.43} \\
    MaPLe & 82.28 & \underline{75.14} & \underline{78.55} \\
    \rowcolor{lightblue}
    \textbf{\textit{This paper}} & \underline{82.36} & \textbf{76.30} & \textbf{79.21} \\
    \bottomrule
    \end{tabular}}
    \end{subtable}
    ~
    \vspace{1em}
    \begin{subtable}[t]{.23\textwidth}
    \centering
    \caption{ImageNet.}
    \scalebox{.85}{
    \begin{tabular}{lcc|c}
    \toprule
    & Base & New & H \\
    \midrule
    CLIP & 72.43 & 68.14 & 70.22 \\
    CoOp & {76.47} & 67.88 & 71.92\\
    CoCoOp & 75.98 & 70.43 & {73.10} \\
    BPL & - & \underline{70.93} & - \\
    MaPLe & \textbf{76.66}  & 70.54  & \underline{73.47}  \\
    \rowcolor{lightblue}
    \textbf{\textit{This paper}} & \underline{76.63} & \textbf{71.33} & \textbf{73.88} \\
    \bottomrule
    \end{tabular}}
    \end{subtable}
    ~
    \begin{subtable}[t]{.23\textwidth}
    \centering
    \caption{Caltech101.}
    \scalebox{.85}{
    \begin{tabular}{lcc|c}
    \toprule
    & Base & New & H \\
    \midrule
    CLIP & 96.84 & 94.00 & 95.40 \\
    CoOp & \underline{98.00} & 89.81 & 93.73 \\
    CoCoOp & 97.96 & 93.81 & {95.84} \\
    BPL & - & \textbf{94.93} & - \\
    MaPLe & 97.74 & \underline{94.36} & \underline{96.02} \\
    \rowcolor{lightblue}
    \textbf{\textit{This paper}} & \textbf{98.28} & {94.27} & \textbf{96.23} \\
    \bottomrule
    \end{tabular}}
    \end{subtable}
    ~
    \begin{subtable}[t]{.23\textwidth}
    \centering
    \caption{OxfordPets.}
    \scalebox{.85}{
    \begin{tabular}{lcc|c}
    \toprule
    & Base & New & H \\
    \midrule
    CLIP & 91.17 & 97.26 & 94.12 \\
    CoOp & 93.67 & 95.29 & 94.47 \\
    CoCoOp & {95.20} & 97.69 & {96.43} \\
    BPL & - & \textbf{98.00} & - \\
    MaPLe & \underline{95.43} & 97.76 & \underline{96.58} \\
    \rowcolor{lightblue}
    \textbf{\textit{This paper}} & \textbf{95.78} & \underline{97.80} & \textbf{96.78} \\
    \bottomrule
    \end{tabular}}
    \end{subtable}
    ~
    \vspace{1mm}
    \begin{subtable}[t]{.23\textwidth}
    \centering
    \caption{StanfordCars.}
    \scalebox{.85}{
    \begin{tabular}{lcc|c}
    \toprule
    & Base & New & H \\
    \midrule
    CLIP & 63.37 & 74.89 & 68.65 \\
    CoOp & \textbf{78.12} & 60.40 & 68.13 \\
    CoCoOp & 70.49 & {73.59} & {72.01} \\
    BPL & - & 73.23 & - \\
    MaPLe & 72.94 & \underline{74.00} & \underline{73.47} \\
    \rowcolor{lightblue}
    \textbf{\textit{This paper}} & \underline{73.05} & \textbf{75.83} &  \textbf{74.41} \\
    \bottomrule
    \end{tabular}}
    \end{subtable}
    ~
    \begin{subtable}[t]{.23\textwidth}
    \centering
    \caption{Flowers102.}
    \scalebox{.85}{
    \begin{tabular}{lcc|c}
    \toprule
    & Base & New & H \\
    \midrule
    CLIP & 72.08 & \textbf{77.80} & 74.83 \\
    CoOp & \textbf{97.60} & 59.67 & 74.06 \\
    CoCoOp & 94.87 & 71.75 & {81.71} \\
    BPL & - & 70.40 & - \\
    MaPLe & 95.92 & 72.46 & \underline{82.56}\\
    \rowcolor{lightblue}
    \textbf{\textit{This paper}} & \underline{96.50} & \underline{76.20} & \textbf{85.16} \\
    \bottomrule
    \end{tabular}}
    \end{subtable}
    ~
    \begin{subtable}[t]{.23\textwidth}
    \centering
    \caption{Food101.}
    \scalebox{.85}{
    \begin{tabular}{lcc|c}
    \toprule
    & Base & New & H \\
    \midrule
    CLIP & 90.10 & 91.22 & 90.66 \\
    CoOp & 88.33 & 82.26 & 85.19 \\
    CoCoOp & {90.70} & 91.29 & {90.99} \\
    BPL & - & \textbf{92.13} & - \\
    MaPLe & \underline{90.71} & \underline{92.05} & \textbf{91.38}\\
    \rowcolor{lightblue}
    \textbf{\textit{This paper}} & \textbf{90.87} & {91.35} & \underline{91.11} \\
    \bottomrule
    \end{tabular}}
    \end{subtable}
    ~
    \begin{subtable}[t]{.23\textwidth}
    \centering
    \caption{FGVCAircraft.}
    \scalebox{.85}{
    \begin{tabular}{lcc|c}
    \toprule
    & Base & New & H \\
    \midrule
    CLIP & 27.19 & \textbf{36.29} & {31.09} \\
    CoOp & \textbf{40.44} & 22.30 & 28.75 \\
    CoCoOp & 33.41 & 23.71 & 27.74 \\
    BPL & - & 35.00 & - \\
    MaPLe & \underline{37.44} & 35.61 & \textbf{36.50}\\
    \rowcolor{lightblue}
    \textbf{\textit{This paper}} & {37.10} & \underline{35.70} & \underline{36.39} \\
    \bottomrule
    \end{tabular}}
    \end{subtable}
    \vspace{1em}
    ~
    \begin{subtable}[t]{.23\textwidth}
    \centering
    \caption{SUN397.}
    \scalebox{.85}{
    \begin{tabular}{lcc|c}
    \toprule
    & Base & New & H \\
    \midrule
    CLIP & 69.36 & 75.35 & 72.23 \\
    CoOp & \underline{80.60} & 65.89 & 72.51 \\
    CoCoOp & 79.74 & 76.86 & {78.27} \\
    BPL & - & {77.87} & - \\
    MaPLe & \textbf{80.82} & \textbf{78.70} & \textbf{79.75}\\
    \rowcolor{lightblue}
    \textbf{\textit{This paper}} & {80.50} & \underline{78.50} & \underline{79.48} \\
    \bottomrule
    \end{tabular}}
    \end{subtable}
    ~
    \begin{subtable}[t]{.23\textwidth}
    \centering
    \caption{DTD.}
    \scalebox{.85}{
    \begin{tabular}{lcc|c}
    \toprule
    & Base & New & H \\
    \midrule
    CLIP & 53.24 & 59.90 & 56.37 \\
    CoOp & {79.44} & 41.18 & 54.24 \\
    CoCoOp & 77.01 & 56.00 &  {64.85} \\
    BPL & - & \underline{60.80} & - \\
    MaPLe & \textbf{80.36} & 59.18 & \underline{68.16} \\
    \rowcolor{lightblue}
    \textbf{\textit{This paper}} & \underline{79.63} & \textbf{61.98} & \textbf{69.71} \\
    \bottomrule
    \end{tabular}}
    \end{subtable}
    ~
    \begin{subtable}[t]{.23\textwidth}
    \centering
    \caption{EuroSAT.}
    \scalebox{.85}{
    \begin{tabular}{lcc|c}
    \toprule
    & Base & New & H \\
    \midrule
    CLIP & 56.48 & 64.05 & 60.03 \\
    CoOp & {92.19} & 54.74 & 68.69 \\
    CoCoOp & 87.49 & 60.04 & \underline{71.21} \\
    BPL & - & \underline{75.30} & - \\
    MaPLe & \textbf{94.07} & 73.23 & \underline{82.35} \\
    \rowcolor{lightblue}
    \textbf{\textit{This paper}} & \underline{93.07} & \textbf{77.63} & \textbf{84.65} \\
    \bottomrule
    \end{tabular}}
    \end{subtable}
    ~
    \begin{subtable}[t]{.23\textwidth}
    \centering
    \caption{UCF101.}
    \scalebox{.85}{
    \begin{tabular}{lcc|c}
    \toprule
    & Base & New & H \\
    \midrule
    CLIP & 70.53 & {77.50} & 73.85 \\
    CoOp & \textbf{84.69} & 56.05 & 67.46 \\
    CoCoOp & 82.33 & 73.45 & {77.64} \\
    BPL & - & 75.77 & - \\
    MaPLe & 83.00 & \underline{78.66} & \underline{80.77} \\
    \rowcolor{lightblue}
    \textbf{\textit{This paper}} & \underline{84.60} & \textbf{78.70} & \textbf{81.54} \\
    \bottomrule
    \end{tabular}}
    \end{subtable}
    \vspace{-5mm}
    \caption{\textbf{Label shift comparison}. The models are trained on the base classes with 16 shots and evaluated on both the base and new classes. 
    We bold the \textbf{best} results and underline the \underline{runner-up}. H denotes the Harmonic mean \cite{xian2018zero}.
    Our method performs well on both base and new classes, therefore achieving the best overall Harmonic mean, demonstrating the generalization ability across label shifts. 
    }
    \label{tab:labelshift}
    \vspace{-5mm}
\end{table*}

\vspace{1mm}
\noindent
\textbf{Label shift.}
We conduct the experiments on label shift following the base-to-new class generalization setting in Zhou \etal \cite{zhou2022conditional}.
The results on eleven datasets and the averaged performance are provided in Table~\ref{tab:labelshift}.
Since our any-shift prompts encode both training and test information, as well as their relationships, it performs well in both base and new classes, therefore achieving the best overall Harmonic mean on the eleven datasets.
Compared with the original CLIP model, the proposed method achieves better performance in the base classes, showing good adaptation to the downstream tasks with the training information.
Compared with the other prompt learning methods CoOp \cite{zhou2022learning}, CoCoOp \cite{zhou2022conditional}, BPL \cite{derakhshani2023bayesian}, and MaPLe \cite{khattak2023maple}, our method performs best in the new classes on seven of the eleven datasets and is competitive on the other four.
This demonstrates the ability of the method to handle label shift by incorporating the distribution information and their relationships.

\vspace{1mm}
\noindent
\textbf{Concept shift.}
For concept shift, we conduct experiments on the introduced \texttt{ImageNet-Superclass} dataset, where the same images are assigned with different annotations. To do so, we evaluate the ImageNet-trained model on the validation set with the superclass annotations. 
As shown in Table~\ref{tab:conshift}, the prompt learning methods achieve similar performance compared with the original CLIP. 
By contrast, our method improves the performance of CLIP by about 2\%, indicating the ability to handle concept shift.

\begin{table}[t]

    \tabstyle{7pt}
    	\resizebox{0.95\columnwidth}{!}{%
    \begin{tabular}{l l ll}
    \toprule
     & \textbf{Concept Shift} & \multicolumn{2}{l}{\textbf{Conditional Shift}} \\ 
    \cmidrule(lr){2-2} \cmidrule(lr){3-4}
    Method & {ImageNet-Superclass} & {Living-17} & {Entity-30} \\
    \midrule
    CLIP\dag & 69.23 & 86.94 & 67.95\\
    CoOp\dag & 69.35 & 87.11 &  78.02   \\
    CoCoOp\dag & 69.77 & 87.24 & 79.52  \\
    \rowcolor{lightblue}
    \textit{\textbf{This paper}} &  \textbf{71.12} \scriptsize{$\pm$ 0.6}  & \textbf{88.41} \scriptsize{$\pm$ 0.3} & \textbf{81.74} \scriptsize{$\pm$ 0.4}  \\
    \bottomrule
    \end{tabular}
    }
    \vspace{-2mm}
\caption{\textbf{Concept shift and conditional shift comparison}. 
The results of the compared methods are based on the author-provided code since the prompt learning methods do not provide results on these shifts.}
\label{tab:conshift}
\vspace{-6mm}
\end{table}

\vspace{1mm}
\noindent
\textbf{Conditional shift.}
We also conduct experiments on two datasets with conditional shift. 
The results are also reported in Table~\ref{tab:conshift}.
The prompt learning methods perform similarly to CLIP while achieving more improvement on \texttt{Entity-30}. 
The reason can be that the class names of \texttt{Living-17} (e.g., wolf, fox) are more detailed than \texttt{Entity-30} (e.g., crustacean, carnivore, insect), revealing the importance of adapting the original CLIP model to downstream tasks in specific scenarios.
Moreover, compared with the conventional prompt learning methods CoOp and CoCoOp, our method consistently improves the performance on both datasets and performs better, demonstrating the effectiveness of any-shift prompting for the conditional shift.

\begin{table}[t]
\begin{center}
	\resizebox{1\columnwidth}{!}{%
\begin{tabular}{llllll}
\toprule
 Method   & Art & Clipart & Product & Real &\textit{\textbf{Mean}} \\ \midrule
CLIP\dag & 79.32 & 67.70 & 86.93 & 87.46 & 80.35\\
CLIP-D\dag & 80.47 & 68.83 & 87.93 & 88.80 & 81.51 \\
CoOp\dag & 80.50 & 69.05 & 88.26 & 89.01 & 81.71 \\
CoCoOp\dag & 80.93 & 69.51 & 88.85  & 89.32 & 82.19 \\
\rowcolor{lightblue}
\textit{\textbf{This paper}}  & \textbf{83.40} \scriptsize{$\pm$ 0.8} & \textbf{72.53} \scriptsize{$\pm$ 0.5} & \textbf{91.24} \scriptsize{$\pm$ 0.6} &  \textbf{90.84} \scriptsize{$\pm$ 0.3} & \textbf{84.50} \scriptsize{$\pm$ 0.4}\\ 
\bottomrule
\end{tabular}
}
\end{center}
\vspace{-5mm}
\caption{\textbf{Multiple shifts comparison} on \texttt{Open-Office-Home}, including both covariate and label shifts.
The results of other methods are based on the author-provided code.}
\label{opendg}
\vspace{-6mm}
\end{table}
\vspace{1mm}
\noindent
\textbf{Joint distribution shift.}
In Table~\ref{opendg}, we report the results on \texttt{Open-Office-Home} for the joint distribution shifts.
Following Shu \etal \cite{shu2021open}, we assign data from different parts of classes in the training domains and evaluate the model on the test domain with both seen and unseen classes. Therefore, the model encounters covariate and label shifts jointly.
As shown in Table~\ref{opendg}, the CLIP-based zero-shot methods keep the same performance as the close-set generalization setting (Table~\ref{tab:dg}) since they are kept frozen.
The prompt learning methods perform slightly worse than the close-set setting. Our method outperforms the others on all test domains, showing the ability to handle joint distribution shifts.

\vspace{1mm}
\noindent 
Overall, our method achieves good performance on covariate, label, concept, conditional, and even joint shifts, demonstrating the effectiveness of handling various distribution shifts by considering the distribution information and their relationship with any-shift prompting.

\subsection{Ablation studies}

\begin{figure}[t]
\centering
\includegraphics[width=0.99\linewidth]{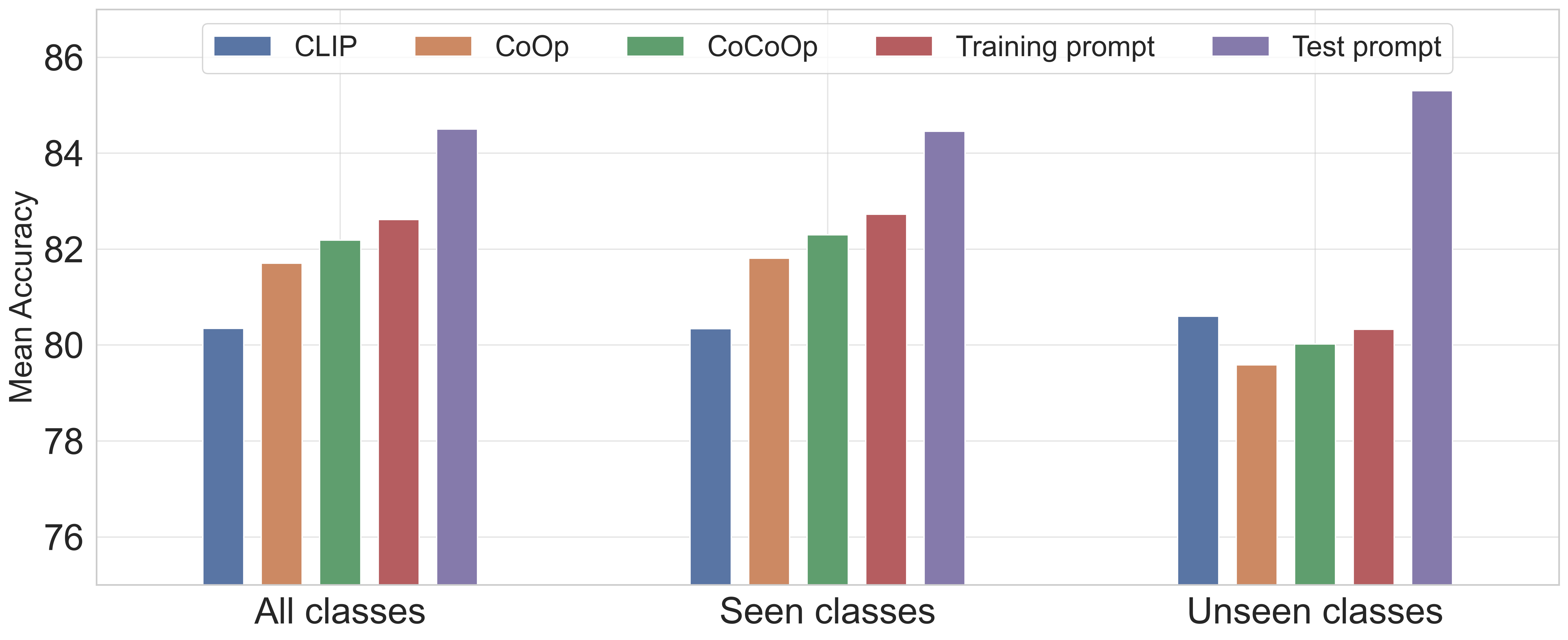}
\vspace{-2mm}
\caption{\textbf{Effectiveness of training and test prompts.} The test prompt in the proposed any-shift prompting achieves good generalization on both seen and unseen classes, indicating its ability to handle different shifts jointly. 
}
\label{fig: vsvt}
\vspace{-2mm}
\end{figure}

\noindent
\textbf{Effectiveness of training and test prompts.}
To investigate the benefits of the training and test prompts of any-shift prompting, we evaluate our method with training and test prompts separately.
The experiments are conducted on \texttt{Open-office-Home} with joint distribution shift. 
We compare the prompts with the original CLIP model as well as CoOp and CoCoOp in Figure~\ref{fig: vsvt}, and provide the accuracy on all classes, seen classes, and unseen classes, respectively. 
CoOp and CoCoOp show better performance on seen classes across covariate shift but struggle in the unseen classes where both covariate shift and label shift exist.  
The training prompt in our method encounters the same problem since it encodes the training information with seen classes but also tends to overfit the training distribution.
The performance is slightly better since it considers uncertainty in the prompt.
By contrast, the test prompt in our method encodes the test information with the relationships between the training and test distribution.
This enables the method to achieve good generalization across different shifts, leading to higher performance on both seen (covariate shift) and unseen classes (both covariate shift and label shift).

\begin{figure}[t]
\centering
\includegraphics[width=0.99\linewidth]{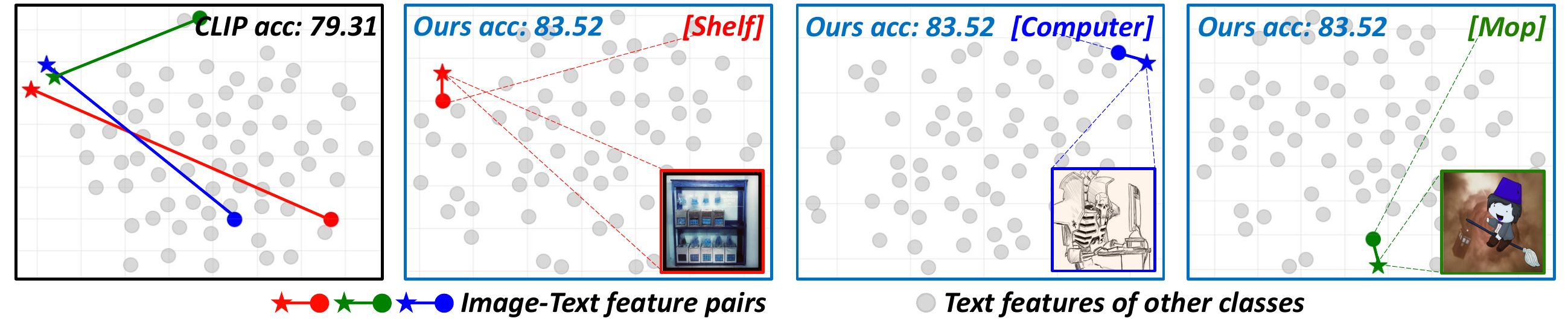}
\vspace{-3mm}
\caption{\textbf{Visualization of generalization effect} on the image and text features before and after generalization. Different colors denote different classes. The image and text features with the same categories get closer after generalization by our method, leading to more accurate predictions.}
\label{fig: tsne_visual}
\vspace{-5mm}
\end{figure}


\begin{table}[t]
\begin{center}
	\resizebox{1\columnwidth}{!}{%
\begin{tabular}{ccc|c}
\toprule
 Training prompt $\bv_s$ & Test text feature of $\mathcal{Y}_t $ & Test image feature of $\x_t$ & Accuracy \\ \midrule
\ymark & & & {82.62} \\
& \ymark & & {82.67} \\ 
& & \ymark & {83.11} \\ 
& \ymark & \ymark & {83.63} \\ 
\ymark & \ymark & \ymark & \textbf{84.50} \\ 
\bottomrule
\end{tabular}
}
\vspace{-2mm}
\caption{\textbf{Benefits of training and test information in any-shift prompt.} The experiments are conducted across the joint shifts on \texttt{Open-Office-Home}. 
Both training and test information in the prompt benefit the method across joint shifts.
}
\label{ablation1}
\end{center}
\vspace{-7mm}
\end{table}


\vspace{1mm}
\noindent
\textbf{Visualization of generalization effect.}
To further show the benefits of generalization with our method, we visualize the image and text features before and after generalization by any-shift prompting.
The experiments are conducted on the ``Art" domain under \texttt{Open-Office-Home}.
The image and text features before generalization are generated by the fixed CLIP image and language encoders respectively.
As shown in Figure~\ref{fig: tsne_visual}, after generalization by any-shift prompting, the image features get closer to the text features of the corresponding ground truth labels, which leads to more accurate predictions.

\vspace{1mm}
\noindent
\textbf{Benefits of training and test information in any-shift prompt.}
To show the benefits of considering different information in the test prompt, we conduct experiments on \texttt{Open-Office-Home}, which contains both covariate and label shifts.
As shown in Table~\ref{ablation1}, using only the training prompt achieves better performance than CLIP (80.35) and we get similar results with only test text features or test image features.
The information from the test images gains more improvement. The reason can be that test images include more unseen information in this setting.
The test prompt generated by both image and text information further improves the generalization of test distributions, indicating the importance of considering test information for generalization.
Moreover, including the training prompt provides the relationships and shift information between training and test distribution in the prompt, leading to the best performance.

\section{Conclusion}

We propose any-shift prompting to adapt the large image-language model (CLIP) to downstream tasks while enhancing the generalization ability across different distribution shifts at test time.
The proposed method bridges the training and test distributions under a hierarchical probabilistic framework, which generates the specific prompt for each test sample by encoding the distribution information and relationships of the training and test distributions.
Once trained, we generate the test-specific prompt across any distribution shift in a single feedforward pass without any fine-tuning or backpropagation.
The test prompt generalizes both the image and language encoders of CLIP to the specific test distribution.
Experiments on various distribution shifts, including covariate shift, label shift, conditional shift, concept shift, and joint shift, demonstrate the effectiveness of the proposed method on the generalization of any test distribution.

\section*{Acknowledgment}
This work is financially supported by the Inception Institute of Artificial Intelligence, the University of Amsterdam, and the allowance 
Top consortia for Knowledge and Innovation (TKIs) from the Netherlands Ministry of Economic Affairs and Climate Policy.

{
    \small
    \bibliographystyle{ieeenat_fullname.bst}
    \bibliography{main.bib}

\begin{thebibliography}{88}
\providecommand{\natexlab}[1]{#1}
\providecommand{\url}[1]{\texttt{#1}}
\expandafter\ifx\csname urlstyle\endcsname\relax
  \providecommand{\doi}[1]{doi: #1}\else
  \providecommand{\doi}{doi: \begingroup \urlstyle{rm}\Url}\fi

\bibitem[Arjovsky et~al.(2019)Arjovsky, Bottou, Gulrajani, and Lopez-Paz]{arjovsky2019invariant}
Martin Arjovsky, L{\'e}on Bottou, Ishaan Gulrajani, and David Lopez-Paz.
\newblock Invariant risk minimization.
\newblock \emph{arXiv preprint arXiv:1907.02893}, 2019.

\bibitem[Azizzadenesheli et~al.(2019)Azizzadenesheli, Liu, Yang, and Anandkumar]{azizzadenesheli2019regularized}
Kamyar Azizzadenesheli, Anqi Liu, Fanny Yang, and Animashree Anandkumar.
\newblock Regularized learning for domain adaptation under label shifts.
\newblock \emph{arXiv preprint arXiv:1903.09734}, 2019.

\bibitem[Bahng et~al.(2022)Bahng, Jahanian, Sankaranarayanan, and Isola]{bahng2022exploring}
Hyojin Bahng, Ali Jahanian, Swami Sankaranarayanan, and Phillip Isola.
\newblock Exploring visual prompts for adapting large-scale models.
\newblock \emph{arXiv preprint arXiv:2203.17274}, 2022.

\bibitem[Balaji et~al.(2018)Balaji, Sankaranarayanan, and Chellappa]{balaji2018metareg}
Yogesh Balaji, Swami Sankaranarayanan, and Rama Chellappa.
\newblock {MetaReg}: Towards domain generalization using meta-regularization.
\newblock In \emph{Advances in Neural Information Processing Systems}, pages 998--1008, 2018.

\bibitem[Bossard et~al.(2014)Bossard, Guillaumin, and Van~Gool]{bossard2014food}
Lukas Bossard, Matthieu Guillaumin, and Luc Van~Gool.
\newblock Food-101--mining discriminative components with random forests.
\newblock In \emph{European Conference on Computer Vision}, pages 446--461. Springer, 2014.

\bibitem[Choi et~al.(2010)Choi, Lim, Torralba, and Willsky]{choi2010exploiting}
Myung~Jin Choi, Joseph~J Lim, Antonio Torralba, and Alan~S Willsky.
\newblock Exploiting hierarchical context on a large database of object categories.
\newblock In \emph{2010 IEEE Computer Society Conference on Computer Vision and Pattern Recognition}, pages 129--136. IEEE, 2010.

\bibitem[Cimpoi et~al.(2014)Cimpoi, Maji, Kokkinos, Mohamed, and Vedaldi]{cimpoi2014describing}
Mircea Cimpoi, Subhransu Maji, Iasonas Kokkinos, Sammy Mohamed, and Andrea Vedaldi.
\newblock Describing textures in the wild.
\newblock In \emph{IEEE Conference on Computer Vision and Pattern Recognition}, pages 3606--3613, 2014.

\bibitem[Deng et~al.(2009)Deng, Dong, Socher, Li, Li, and Fei-Fei]{deng2009imagenet}
Jia Deng, Wei Dong, Richard Socher, Li-Jia Li, Kai Li, and Li Fei-Fei.
\newblock {ImageNet}: A large-scale hierarchical image database.
\newblock In \emph{IEEE Conference on Computer Vision and Pattern Recognition}, pages 248--255, 2009.

\bibitem[Derakhshani et~al.(2023)Derakhshani, Sanchez, Bulat, da~Costa, Snoek, Tzimiropoulos, and Martinez]{derakhshani2023bayesian}
Mohammad~Mahdi Derakhshani, Enrique Sanchez, Adrian Bulat, Victor G~Turrisi da Costa, Cees~GM Snoek, Georgios Tzimiropoulos, and Brais Martinez.
\newblock Bayesian prompt learning for image-language model generalization.
\newblock In \emph{IEEE International Conference on Computer Vision}, pages 15237--15246, 2023.

\bibitem[Dosovitskiy et~al.(2020)Dosovitskiy, Beyer, Kolesnikov, Weissenborn, Zhai, Unterthiner, Dehghani, Minderer, Heigold, Gelly, et~al.]{dosovitskiy2020image}
Alexey Dosovitskiy, Lucas Beyer, Alexander Kolesnikov, Dirk Weissenborn, Xiaohua Zhai, Thomas Unterthiner, Mostafa Dehghani, Matthias Minderer, Georg Heigold, Sylvain Gelly, et~al.
\newblock An image is worth 16x16 words: Transformers for image recognition at scale.
\newblock In \emph{International Conference on Learning Representations}, 2020.

\bibitem[Dou et~al.(2019)Dou, Castro, Kamnitsas, and Glocker]{dou2019domain}
Qi Dou, Daniel~C Castro, Konstantinos Kamnitsas, and Ben Glocker.
\newblock Domain generalization via model-agnostic learning of semantic features.
\newblock In \emph{Advances in Neural Information Processing Systems}, 2019.

\bibitem[Dubey et~al.(2021)Dubey, Ramanathan, Pentland, and Mahajan]{dubey2021adaptive}
Abhimanyu Dubey, Vignesh Ramanathan, Alex Pentland, and Dhruv Mahajan.
\newblock Adaptive methods for real-world domain generalization.
\newblock In \emph{IEEE Conference on Computer Vision and Pattern Recognition}, pages 14340--14349, 2021.

\bibitem[Everingham et~al.(2010)Everingham, Van~Gool, Williams, Winn, and Zisserman]{everingham2010pascal}
Mark Everingham, Luc Van~Gool, Christopher~KI Williams, John Winn, and Andrew Zisserman.
\newblock The pascal visual object classes (voc) challenge.
\newblock \emph{International Journal of Computer Vision}, 88\penalty0 (2):\penalty0 303--338, 2010.

\bibitem[Fang et~al.(2013)Fang, Lin, Chen, Tsai, and Lin]{fang2013video}
Yuming Fang, Weisi Lin, Zhenzhong Chen, Chia-Ming Tsai, and Chia-Wen Lin.
\newblock A video saliency detection model in compressed domain.
\newblock \emph{IEEE Transactions on Circuits and Systems for Video Technology}, 24\penalty0 (1):\penalty0 27--38, 2013.

\bibitem[Fei-Fei et~al.(2004)Fei-Fei, Fergus, and Perona]{fei2004learning}
Li Fei-Fei, Rob Fergus, and Pietro Perona.
\newblock Learning generative visual models from few training examples: An incremental bayesian approach tested on 101 object categories.
\newblock In \emph{IEEE Conference on Computer Vision and Pattern Recognition Workshop}, pages 178--178. IEEE, 2004.

\bibitem[Gao et~al.(2023)Gao, Geng, Zhang, Ma, Fang, Zhang, Li, and Qiao]{gao2023clip}
Peng Gao, Shijie Geng, Renrui Zhang, Teli Ma, Rongyao Fang, Yongfeng Zhang, Hongsheng Li, and Yu Qiao.
\newblock Clip-adapter: Better vision-language models with feature adapters.
\newblock \emph{International Journal of Computer Vision}, pages 1--15, 2023.

\bibitem[Garg et~al.(2023)Garg, Erickson, Sharpnack, Smola, Balakrishnan, and Lipton]{garg2023rlsbench}
Saurabh Garg, Nick Erickson, James Sharpnack, Alex Smola, Sivaraman Balakrishnan, and Zachary~Chase Lipton.
\newblock Rlsbench: Domain adaptation under relaxed label shift.
\newblock In \emph{International Conference on Machine Learning}, pages 10879--10928. PMLR, 2023.

\bibitem[Gong et~al.(2016)Gong, Zhang, Liu, Tao, Glymour, and Sch{\"o}lkopf]{gong2016domain}
Mingming Gong, Kun Zhang, Tongliang Liu, Dacheng Tao, Clark Glymour, and Bernhard Sch{\"o}lkopf.
\newblock Domain adaptation with conditional transferable components.
\newblock In \emph{International Conference on Machine Learning}, pages 2839--2848. PMLR, 2016.

\bibitem[Goyal et~al.(2022)Goyal, Sun, Raghunathan, and Kolter]{goyal2022test}
Sachin Goyal, Mingjie Sun, Aditi Raghunathan, and Zico Kolter.
\newblock Test-time adaptation via conjugate pseudo-labels.
\newblock In \emph{Advances in Neural Information Processing Systems}, 2022.

\bibitem[Griffin et~al.(2007)Griffin, Holub, and Perona]{griffin2007caltech}
Gregory Griffin, Alex Holub, and Pietro Perona.
\newblock Caltech-256 object category dataset.
\newblock 2007.

\bibitem[Gulrajani and Lopez-Paz(2020)]{gulrajani2020search}
Ishaan Gulrajani and David Lopez-Paz.
\newblock In search of lost domain generalization.
\newblock In \emph{International Conference on Learning Representations}, 2020.

\bibitem[Guo et~al.(2020)Guo, Gong, Liu, Zhang, and Tao]{guo2020ltf}
Jiaxian Guo, Mingming Gong, Tongliang Liu, Kun Zhang, and Dacheng Tao.
\newblock Ltf: A label transformation framework for correcting label shift.
\newblock In \emph{International Conference on Machine Learning}, pages 3843--3853. PMLR, 2020.

\bibitem[Helber et~al.(2019)Helber, Bischke, Dengel, and Borth]{helber2019eurosat}
Patrick Helber, Benjamin Bischke, Andreas Dengel, and Damian Borth.
\newblock Eurosat: A novel dataset and deep learning benchmark for land use and land cover classification.
\newblock \emph{IEEE Journal of Selected Topics in Applied Earth Observations and Remote Sensing}, 12\penalty0 (7):\penalty0 2217--2226, 2019.

\bibitem[Hendrycks and Dietterich(2019)]{hendrycks2019benchmarking}
Dan Hendrycks and Thomas Dietterich.
\newblock Benchmarking neural network robustness to common corruptions and perturbations.
\newblock \emph{arXiv preprint arXiv:1903.12261}, 2019.

\bibitem[Hendrycks et~al.(2021{\natexlab{a}})Hendrycks, Basart, Mu, Kadavath, Wang, Dorundo, Desai, Zhu, Parajuli, Guo, et~al.]{hendrycks2021many}
Dan Hendrycks, Steven Basart, Norman Mu, Saurav Kadavath, Frank Wang, Evan Dorundo, Rahul Desai, Tyler Zhu, Samyak Parajuli, Mike Guo, et~al.
\newblock The many faces of robustness: A critical analysis of out-of-distribution generalization.
\newblock In \emph{IEEE International Conference on Computer Vision}, pages 8340--8349, 2021{\natexlab{a}}.

\bibitem[Hendrycks et~al.(2021{\natexlab{b}})Hendrycks, Zhao, Basart, Steinhardt, and Song]{hendrycks2021natural}
Dan Hendrycks, Kevin Zhao, Steven Basart, Jacob Steinhardt, and Dawn Song.
\newblock Natural adversarial examples.
\newblock In \emph{IEEE Conference on Computer Vision and Pattern Recognition}, pages 15262--15271, 2021{\natexlab{b}}.

\bibitem[Iwasawa et~al.(2021)]{iwasawa2021test}
Yusuke Iwasawa et~al.
\newblock Test-time classifier adjustment module for model-agnostic domain generalization.
\newblock In \emph{Advances in Neural Information Processing Systems}, 2021.

\bibitem[Jia et~al.(2021)Jia, Yang, Xia, Chen, Parekh, Pham, Le, Sung, Li, and Duerig]{jia2021scaling}
Chao Jia, Yinfei Yang, Ye Xia, Yi-Ting Chen, Zarana Parekh, Hieu Pham, Quoc Le, Yun-Hsuan Sung, Zhen Li, and Tom Duerig.
\newblock Scaling up visual and vision-language representation learning with noisy text supervision.
\newblock In \emph{International Conference on Machine Learning}, pages 4904--4916. PMLR, 2021.

\bibitem[Khattak et~al.(2023)Khattak, Rasheed, Maaz, Khan, and Khan]{khattak2023maple}
Muhammad~Uzair Khattak, Hanoona Rasheed, Muhammad Maaz, Salman Khan, and Fahad~Shahbaz Khan.
\newblock Maple: Multi-modal prompt learning.
\newblock In \emph{IEEE Conference on Computer Vision and Pattern Recognition}, pages 19113--19122, 2023.

\bibitem[Kingma and Welling(2013)]{kingma2013auto}
Diederik~P Kingma and Max Welling.
\newblock Auto-encoding variational bayes.
\newblock \emph{arXiv preprint arXiv:1312.6114}, 2013.

\bibitem[Krause et~al.(2013)Krause, Stark, Deng, and Fei-Fei]{krause20133d}
Jonathan Krause, Michael Stark, Jia Deng, and Li Fei-Fei.
\newblock 3d object representations for fine-grained categorization.
\newblock In \emph{IEEE International Conference on Computer Vision Workshops}, pages 554--561, 2013.

\bibitem[Lee et~al.(2023)Lee, Chen, Tajwar, Kumar, Yao, Liang, and Finn]{lee2022surgical}
Yoonho Lee, Annie~S Chen, Fahim Tajwar, Ananya Kumar, Huaxiu Yao, Percy Liang, and Chelsea Finn.
\newblock Surgical fine-tuning improves adaptation to distribution shifts.
\newblock In \emph{International Conference on Learning Representations}, 2023.

\bibitem[Lester et~al.(2021)Lester, Al-Rfou, and Constant]{lester2021power}
Brian Lester, Rami Al-Rfou, and Noah Constant.
\newblock The power of scale for parameter-efficient prompt tuning.
\newblock \emph{arXiv preprint arXiv:2104.08691}, 2021.

\bibitem[Li et~al.(2017)Li, Yang, Song, and Hospedales]{li2017deeper}
Da Li, Yongxin Yang, Yi-Zhe Song, and Timothy~M Hospedales.
\newblock Deeper, broader and artier domain generalization.
\newblock In \emph{IEEE International Conference on Computer Vision}, pages 5542--5550, 2017.

\bibitem[Li et~al.(2018{\natexlab{a}})Li, Yang, Song, and Hospedales]{li2018metalearning}
Da Li, Yongxin Yang, Yi-Zhe Song, and Timothy Hospedales.
\newblock Learning to generalize: Meta-learning for domain generalization.
\newblock In \emph{AAAI Conference on Artificial Intelligence}, 2018{\natexlab{a}}.

\bibitem[Li et~al.(2018{\natexlab{b}})Li, Jialin~Pan, Wang, and Kot]{li2018domain}
Haoliang Li, Sinno Jialin~Pan, Shiqi Wang, and Alex~C Kot.
\newblock Domain generalization with adversarial feature learning.
\newblock In \emph{IEEE Conference on Computer Vision and Pattern Recognition}, pages 5400--5409, 2018{\natexlab{b}}.

\bibitem[Li and Liang(2021)]{li2021prefix}
Xiang~Lisa Li and Percy Liang.
\newblock Prefix-tuning: Optimizing continuous prompts for generation.
\newblock \emph{arXiv preprint arXiv:2101.00190}, 2021.

\bibitem[Liang et~al.(2020)Liang, Hu, and Feng]{liang2020we}
Jian Liang, Dapeng Hu, and Jiashi Feng.
\newblock Do we really need to access the source data? source hypothesis transfer for unsupervised domain adaptation.
\newblock In \emph{International Conference on Machine Learning}, pages 6028--6039. PMLR, 2020.

\bibitem[Lim et~al.(2023)Lim, Kim, Choo, and Choi]{lim2023ttn}
Hyesu Lim, Byeonggeun Kim, Jaegul Choo, and Sungha Choi.
\newblock Ttn: A domain-shift aware batch normalization in test-time adaptation.
\newblock In \emph{International Conference on Learning Representations}, 2023.

\bibitem[Liu et~al.(2023)Liu, Yuan, Fu, Jiang, Hayashi, and Neubig]{liu2023pre}
Pengfei Liu, Weizhe Yuan, Jinlan Fu, Zhengbao Jiang, Hiroaki Hayashi, and Graham Neubig.
\newblock Pre-train, prompt, and predict: A systematic survey of prompting methods in natural language processing.
\newblock \emph{ACM Computing Surveys}, 55\penalty0 (9):\penalty0 1--35, 2023.

\bibitem[Liu et~al.(2021{\natexlab{a}})Liu, Guo, Li, Xing, You, Kuo, El~Fakhri, and Woo]{liu2021adversarial}
Xiaofeng Liu, Zhenhua Guo, Site Li, Fangxu Xing, Jane You, C-C~Jay Kuo, Georges El~Fakhri, and Jonghye Woo.
\newblock Adversarial unsupervised domain adaptation with conditional and label shift: Infer, align and iterate.
\newblock In \emph{IEEE International Conference on Computer Vision}, pages 10367--10376, 2021{\natexlab{a}}.

\bibitem[Liu et~al.(2022)Liu, Yoo, Xing, Oh, El~Fakhri, Kang, Woo, et~al.]{liu2022deep}
Xiaofeng Liu, Chaehwa Yoo, Fangxu Xing, Hyejin Oh, Georges El~Fakhri, Je-Won Kang, Jonghye Woo, et~al.
\newblock Deep unsupervised domain adaptation: A review of recent advances and perspectives.
\newblock \emph{APSIPA Transactions on Signal and Information Processing}, 11\penalty0 (1), 2022.

\bibitem[Liu et~al.(2021{\natexlab{b}})Liu, Kothari, van Delft, Bellot-Gurlet, Mordan, and Alahi]{liu2021ttt++}
Yuejiang Liu, Parth Kothari, Bastien van Delft, Baptiste Bellot-Gurlet, Taylor Mordan, and Alexandre Alahi.
\newblock Ttt++: When does self-supervised test-time training fail or thrive?
\newblock In \emph{Advances in Neural Information Processing Systems}, 2021{\natexlab{b}}.

\bibitem[Long et~al.(2015)Long, Cao, Wang, and Jordan]{long2015learning}
Mingsheng Long, Yue Cao, Jianmin Wang, and Michael Jordan.
\newblock Learning transferable features with deep adaptation networks.
\newblock In \emph{International Conference on Machine Learning}, pages 97--105. PMLR, 2015.

\bibitem[Maji et~al.(2013)Maji, Rahtu, Kannala, Blaschko, and Vedaldi]{maji2013fine}
Subhransu Maji, Esa Rahtu, Juho Kannala, Matthew Blaschko, and Andrea Vedaldi.
\newblock Fine-grained visual classification of aircraft.
\newblock \emph{arXiv preprint arXiv:1306.5151}, 2013.

\bibitem[Motiian et~al.(2017)Motiian, Piccirilli, Adjeroh, and Doretto]{motiian2017unified}
Saeid Motiian, Marco Piccirilli, Donald~A Adjeroh, and Gianfranco Doretto.
\newblock Unified deep supervised domain adaptation and generalization.
\newblock In \emph{IEEE International Conference on Computer Vision}, pages 5715--5725, 2017.

\bibitem[Muandet et~al.(2013)Muandet, Balduzzi, and Sch{\"o}lkopf]{muandet2013domain}
Krikamol Muandet, David Balduzzi, and Bernhard Sch{\"o}lkopf.
\newblock Domain generalization via invariant feature representation.
\newblock In \emph{International Conference on Machine Learning}, pages 10--18. PMLR, 2013.

\bibitem[Nilsback and Zisserman(2008)]{nilsback2008automated}
Maria-Elena Nilsback and Andrew Zisserman.
\newblock Automated flower classification over a large number of classes.
\newblock In \emph{2008 Sixth Indian conference on computer vision, graphics \& image processing}, pages 722--729. IEEE, 2008.

\bibitem[Niu et~al.(2022)Niu, Wu, Zhang, Chen, Zheng, Zhao, and Tan]{niu2022efficient}
Shuaicheng Niu, Jiaxiang Wu, Yifan Zhang, Yaofo Chen, Shijian Zheng, Peilin Zhao, and Mingkui Tan.
\newblock Efficient test-time model adaptation without forgetting.
\newblock In \emph{International Conference on Machine Learning}, pages 16888--16905. PMLR, 2022.

\bibitem[Niu et~al.(2023)Niu, Wu, Zhang, Wen, Chen, Zhao, and Tan]{niu2023towards}
Shuaicheng Niu, Jiaxiang Wu, Yifan Zhang, Zhiquan Wen, Yaofo Chen, Peilin Zhao, and Mingkui Tan.
\newblock Towards stable test-time adaptation in dynamic wild world.
\newblock In \emph{International Conference on Learning Representations}, 2023.

\bibitem[Novack et~al.(2023)Novack, McAuley, Lipton, and Garg]{novack2023chils}
Zachary Novack, Julian McAuley, Zachary~Chase Lipton, and Saurabh Garg.
\newblock Chils: Zero-shot image classification with hierarchical label sets.
\newblock In \emph{International Conference on Machine Learning}, pages 26342--26362. PMLR, 2023.

\bibitem[Park et~al.(2023)Park, Yang, Choo, and Yun]{park2023label}
Sunghyun Park, Seunghan Yang, Jaegul Choo, and Sungrack Yun.
\newblock Label shift adapter for test-time adaptation under covariate and label shifts.
\newblock In \emph{IEEE International Conference on Computer Vision}, pages 16421--16431, 2023.

\bibitem[Parkhi et~al.(2012)Parkhi, Vedaldi, Zisserman, and Jawahar]{parkhi2012cats}
Omkar~M Parkhi, Andrea Vedaldi, Andrew Zisserman, and CV Jawahar.
\newblock Cats and dogs.
\newblock In \emph{IEEE Conference on Computer Vision and Pattern Recognition}, pages 3498--3505. IEEE, 2012.

\bibitem[Peng et~al.(2019)Peng, Bai, Xia, Huang, Saenko, and Wang]{peng2019moment}
Xingchao Peng, Qinxun Bai, Xide Xia, Zijun Huang, Kate Saenko, and Bo Wang.
\newblock Moment matching for multi-source domain adaptation.
\newblock In \emph{IEEE International Conference on Computer Vision}, pages 1406--1415, 2019.

\bibitem[Radford et~al.(2021)Radford, Kim, Hallacy, Ramesh, Goh, Agarwal, Sastry, Askell, Mishkin, Clark, et~al.]{radford2021learning}
Alec Radford, Jong~Wook Kim, Chris Hallacy, Aditya Ramesh, Gabriel Goh, Sandhini Agarwal, Girish Sastry, Amanda Askell, Pamela Mishkin, Jack Clark, et~al.
\newblock Learning transferable visual models from natural language supervision.
\newblock In \emph{International Conference on Machine Learning}, pages 8748--8763. PMLR, 2021.

\bibitem[Ramesh et~al.(2022)Ramesh, Dhariwal, Nichol, Chu, and Chen]{ramesh2022hierarchical}
Aditya Ramesh, Prafulla Dhariwal, Alex Nichol, Casey Chu, and Mark Chen.
\newblock Hierarchical text-conditional image generation with clip latents.
\newblock \emph{arXiv preprint arXiv:2204.06125}, 2022.

\bibitem[Recht et~al.(2019)Recht, Roelofs, Schmidt, and Shankar]{recht2019imagenet}
Benjamin Recht, Rebecca Roelofs, Ludwig Schmidt, and Vaishaal Shankar.
\newblock Do imagenet classifiers generalize to imagenet?
\newblock In \emph{International Conference on Machine Learning}, pages 5389--5400. PMLR, 2019.

\bibitem[Roberts et~al.(2022)Roberts, Mani, Garg, and Lipton]{roberts2022unsupervised}
Manley Roberts, Pranav Mani, Saurabh Garg, and Zachary Lipton.
\newblock Unsupervised learning under latent label shift.
\newblock In \emph{Advances in Neural Information Processing Systems}, pages 18763--18778, 2022.

\bibitem[Roth et~al.(2023)Roth, Kim, Koepke, Vinyals, Schmid, and Akata]{roth2023waffling}
Karsten Roth, Jae~Myung Kim, A Koepke, Oriol Vinyals, Cordelia Schmid, and Zeynep Akata.
\newblock Waffling around for performance: Visual classification with random words and broad concepts.
\newblock \emph{arXiv preprint arXiv:2306.07282}, 2023.

\bibitem[Russell et~al.(2008)Russell, Torralba, Murphy, and Freeman]{russell2008labelme}
Bryan~C Russell, Antonio Torralba, Kevin~P Murphy, and William~T Freeman.
\newblock Labelme: a database and web-based tool for image annotation.
\newblock \emph{International Journal of Computer Vision}, 77\penalty0 (1-3):\penalty0 157--173, 2008.

\bibitem[Samadh et~al.(2023)Samadh, Gani, Hussein, Khattak, Naseer, Khan, and Khan]{samadh2023align}
Jameel Hassan~Abdul Samadh, Hanan Gani, Noor~Hazim Hussein, Muhammad~Uzair Khattak, Muzammal Naseer, Fahad Khan, and Salman Khan.
\newblock Align your prompts: Test-time prompting with distribution alignment for zero-shot generalization.
\newblock In \emph{Advances in Neural Information Processing Systems}, 2023.

\bibitem[Santurkar et~al.(2020)Santurkar, Tsipras, and Madry]{santurkar2020breeds}
Shibani Santurkar, Dimitris Tsipras, and Aleksander Madry.
\newblock Breeds: Benchmarks for subpopulation shift.
\newblock \emph{arXiv preprint arXiv:2008.04859}, 2020.

\bibitem[Schneider et~al.(2020)Schneider, Rusak, Eck, Bringmann, Brendel, and Bethge]{schneider2020improving}
Steffen Schneider, Evgenia Rusak, Luisa Eck, Oliver Bringmann, Wieland Brendel, and Matthias Bethge.
\newblock Improving robustness against common corruptions by covariate shift adaptation.
\newblock In \emph{Advances in Neural Information Processing Systems}, pages 11539--11551, 2020.

\bibitem[Shen et~al.(2022)Shen, Xiao, Zhen, Snoek, and Worring]{shen2022association}
Jiayi Shen, Zehao Xiao, Xiantong Zhen, Cees Snoek, and Marcel Worring.
\newblock Association graph learning for multi-task classification with category shifts.
\newblock In \emph{Advances in Neural Information Processing Systems}, pages 4503--4516, 2022.

\bibitem[Shu et~al.(2022)Shu, Nie, Huang, Yu, Goldstein, Anandkumar, and Xiao]{shu2022test}
Manli Shu, Weili Nie, De-An Huang, Zhiding Yu, Tom Goldstein, Anima Anandkumar, and Chaowei Xiao.
\newblock Test-time prompt tuning for zero-shot generalization in vision-language models.
\newblock In \emph{Advances in Neural Information Processing Systems}, pages 14274--14289, 2022.

\bibitem[Shu et~al.(2021)Shu, Cao, Wang, Wang, and Long]{shu2021open}
Yang Shu, Zhangjie Cao, Chenyu Wang, Jianmin Wang, and Mingsheng Long.
\newblock Open domain generalization with domain-augmented meta-learning.
\newblock In \emph{IEEE Conference on Computer Vision and Pattern Recognition}, pages 9624--9633, 2021.

\bibitem[Soomro et~al.(2012)Soomro, Zamir, and Shah]{soomro2012ucf101}
Khurram Soomro, Amir~Roshan Zamir, and Mubarak Shah.
\newblock Ucf101: A dataset of 101 human actions classes from videos in the wild.
\newblock \emph{arXiv preprint arXiv:1212.0402}, 2012.

\bibitem[Sun et~al.(2020)Sun, Wang, Liu, Miller, Efros, and Hardt]{sun2020test}
Yu Sun, Xiaolong Wang, Zhuang Liu, John Miller, Alexei Efros, and Moritz Hardt.
\newblock Test-time training with self-supervision for generalization under distribution shifts.
\newblock In \emph{International Conference on Machine Learning}, pages 9229--9248. PMLR, 2020.

\bibitem[Tachet~des Combes et~al.(2020)Tachet~des Combes, Zhao, Wang, and Gordon]{tachet2020domain}
Remi Tachet~des Combes, Han Zhao, Yu-Xiang Wang, and Geoffrey~J Gordon.
\newblock Domain adaptation with conditional distribution matching and generalized label shift.
\newblock In \emph{Advances in Neural Information Processing Systems}, pages 19276--19289, 2020.

\bibitem[Venkateswara et~al.(2017)Venkateswara, Eusebio, Chakraborty, and Panchanathan]{venkateswara2017deep}
Hemanth Venkateswara, Jose Eusebio, Shayok Chakraborty, and Sethuraman Panchanathan.
\newblock Deep hashing network for unsupervised domain adaptation.
\newblock In \emph{IEEE Conference on Computer Vision and Pattern Recognition}, pages 5018--5027, 2017.

\bibitem[Wang et~al.(2021)Wang, Shelhamer, Liu, Olshausen, and Darrell]{wang2021tent}
Dequan Wang, Evan Shelhamer, Shaoteng Liu, Bruno Olshausen, and Trevor Darrell.
\newblock Tent: Fully test-time adaptation by entropy minimization.
\newblock In \emph{International Conference on Learning Representations}, 2021.

\bibitem[Wang et~al.(2019)Wang, Ge, Lipton, and Xing]{wang2019learning}
Haohan Wang, Songwei Ge, Zachary Lipton, and Eric~P Xing.
\newblock Learning robust global representations by penalizing local predictive power.
\newblock In \emph{Advances in Neural Information Processing Systems}, 2019.

\bibitem[Wang and Deng(2018)]{wang2018deep}
Mei Wang and Weihong Deng.
\newblock Deep visual domain adaptation: A survey.
\newblock \emph{Neurocomputing}, 312:\penalty0 135--153, 2018.

\bibitem[Wu et~al.(2021)Wu, Guo, Su, and Weinberger]{wu2021online}
Ruihan Wu, Chuan Guo, Yi Su, and Kilian~Q Weinberger.
\newblock Online adaptation to label distribution shift.
\newblock In \emph{Advances in Neural Information Processing Systems}, pages 11340--11351, 2021.

\bibitem[Xian et~al.(2018)Xian, Lampert, Schiele, and Akata]{xian2018zero}
Yongqin Xian, Christoph~H Lampert, Bernt Schiele, and Zeynep Akata.
\newblock Zero-shot learning—a comprehensive evaluation of the good, the bad and the ugly.
\newblock \emph{IEEE Transactions on Pattern Analysis and Machine Intelligence}, 41\penalty0 (9):\penalty0 2251--2265, 2018.

\bibitem[Xiao et~al.(2010)Xiao, Hays, Ehinger, Oliva, and Torralba]{xiao2010sun}
Jianxiong Xiao, James Hays, Krista~A Ehinger, Aude Oliva, and Antonio Torralba.
\newblock Sun database: Large-scale scene recognition from abbey to zoo.
\newblock In \emph{2010 IEEE Computer Society Conference on Computer Vision and Pattern Recognition}, pages 3485--3492. IEEE, 2010.

\bibitem[Xiao et~al.(2021)Xiao, Shen, Zhen, Shao, and Snoek]{xiao2021bit}
Zehao Xiao, Jiayi Shen, Xiantong Zhen, Ling Shao, and Cees G~M Snoek.
\newblock A bit more bayesian: Domain-invariant learning with uncertainty.
\newblock In \emph{International Conference on Machine Learning}. PMLR, 2021.

\bibitem[Xiao et~al.(2022)Xiao, Zhen, Shao, and Snoek]{xiao2022learning}
Zehao Xiao, Xiantong Zhen, Ling Shao, and Cees G~M Snoek.
\newblock Learning to generalize across domains on single test samples.
\newblock In \emph{International Conference on Learning Representations}, 2022.

\bibitem[Yao et~al.(2023)Yao, Zhang, and Xu]{yao2023visual}
Hantao Yao, Rui Zhang, and Changsheng Xu.
\newblock Visual-language prompt tuning with knowledge-guided context optimization.
\newblock In \emph{IEEE Conference on Computer Vision and Pattern Recognition}, pages 6757--6767, 2023.

\bibitem[Yi et~al.(2023)Yi, Xu, Xu, Li, Pu, Ling, McLeod, and Wang]{yi2023source}
Li Yi, Gezheng Xu, Pengcheng Xu, Jiaqi Li, Ruizhi Pu, Charles Ling, A~Ian McLeod, and Boyu Wang.
\newblock When source-free domain adaptation meets learning with noisy labels.
\newblock In \emph{International Conference on Learning Representations}, 2023.

\bibitem[Zhang et~al.(2013)Zhang, Sch{\"o}lkopf, Muandet, and Wang]{zhang2013domain}
Kun Zhang, Bernhard Sch{\"o}lkopf, Krikamol Muandet, and Zhikun Wang.
\newblock Domain adaptation under target and conditional shift.
\newblock In \emph{International Conference on Machine Learning}, pages 819--827. PMLR, 2013.

\bibitem[Zhang et~al.(2022)Zhang, Levine, and Finn]{zhang2021memo}
Marvin Zhang, Sergey Levine, and Chelsea Finn.
\newblock Memo: Test time robustness via adaptation and augmentation.
\newblock In \emph{Advances in Neural Information Processing Systems}, pages 38629--38642, 2022.

\bibitem[Zhang et~al.(2021)Zhang, Gu, Matsuo, and Iwasawa]{zhang2021domain}
Xin Zhang, Shixiang~Shane Gu, Yutaka Matsuo, and Yusuke Iwasawa.
\newblock Domain prompt learning for efficiently adapting clip to unseen domains.
\newblock \emph{arXiv e-prints}, pages arXiv--2111, 2021.

\bibitem[Zhang et~al.(2023)Zhang, Wang, Jin, Yuan, Zhang, Wang, Jin, and Tan]{zhang2023adanpc}
Yifan Zhang, Xue Wang, Kexin Jin, Kun Yuan, Zhang Zhang, Liang Wang, Rong Jin, and Tieniu Tan.
\newblock Adanpc: Exploring non-parametric classifier for test-time adaptation.
\newblock In \emph{International Conference on Machine Learning}, pages 41647--41676. PMLR, 2023.

\bibitem[Zhou et~al.(2022{\natexlab{a}})Zhou, Liu, Qiao, Xiang, and Loy]{zhou2022domain}
Kaiyang Zhou, Ziwei Liu, Yu Qiao, Tao Xiang, and Chen~Change Loy.
\newblock Domain generalization: A survey.
\newblock \emph{IEEE Transactions on Pattern Analysis and Machine Intelligence}, 2022{\natexlab{a}}.

\bibitem[Zhou et~al.(2022{\natexlab{b}})Zhou, Yang, Loy, and Liu]{zhou2022conditional}
Kaiyang Zhou, Jingkang Yang, Chen~Change Loy, and Ziwei Liu.
\newblock Conditional prompt learning for vision-language models.
\newblock In \emph{IEEE Conference on Computer Vision and Pattern Recognition}, pages 16816--16825, 2022{\natexlab{b}}.

\bibitem[Zhou et~al.(2022{\natexlab{c}})Zhou, Yang, Loy, and Liu]{zhou2022learning}
Kaiyang Zhou, Jingkang Yang, Chen~Change Loy, and Ziwei Liu.
\newblock Learning to prompt for vision-language models.
\newblock \emph{International Journal of Computer Vision}, 130\penalty0 (9):\penalty0 2337--2348, 2022{\natexlab{c}}.

\bibitem[Zhu et~al.(2023)Zhu, Niu, Han, Wu, and Zhang]{zhu2023prompt}
Beier Zhu, Yulei Niu, Yucheng Han, Yue Wu, and Hanwang Zhang.
\newblock Prompt-aligned gradient for prompt tuning.
\newblock In \emph{IEEE International Conference on Computer Vision}, pages 15659--15669, 2023.

\end{thebibliography}
}

\appendix

\section{Derivations of any-shift prompting}
In the main paper, we provide the modeling of our any-shift prompting. Here we provide further derivations of the optimizations of the prior and posterior distributions.

To model the information of training and test distributions and their relationships, we propose any-shift prompting within a hierarchical framework. 
We introduce training and test prompts as latent variables in the hierarchical probabilistic architecture, the prediction function of the CLIP model is then formulated as:
\begin{equation}
\small
\label{eq: prior_predict}
\begin{aligned}
    & p_{\Phi, \theta}(\y_t|\x_t, \mathcal{Y}_t, \mathcal{D}_s) \\
    & = \int \int p(\y_t, \bv_t, \bv_s|\x_t, \mathcal{Y}_t, \x_s, \y_s, \mathcal{Y}_s) d\bv_t d\bv_s \\
    & = \int \int p(\y_t|\x_t, \bv_t, \mathcal{Y}_t) p(\bv_t, \bv_s|\x_t, \mathcal{Y}_t, \mathcal{D}_s) d\bv_t d\bv_s \\
    & = \int \int p_{\Phi}(\y_t|\x_t, \bv_t, \mathcal{Y}_t) p_{\btheta}(\bv_t|\bv_s, \x_t, \mathcal{Y}_t) p(\bv_s|\mathcal{D}_s) d\bv_t d\bv_s,
\end{aligned}
\end{equation}
where the prior distribution of the training and test prompts is factorized as
\begin{equation}
\small
\label{eq: prior}
\begin{aligned}
p(\bv_t, \bv_s|\x_t, \mathcal{Y}_t, \mathcal{D}_s) {=} p_{\btheta}(\bv_t|\bv_s, \x_t, \mathcal{Y}_t) p(\bv_s|\mathcal{D}_s).
\end{aligned}
\end{equation}
$p(\bv_s|\mathcal{D}_s)$ is learned from the training data $\mathcal{D}_s$ sampled from training distribution $p(\x_s, \y_s)$.
$p_{\btheta}(\bv_t|\bv_s, \x_t, \mathcal{Y}_t)$ denotes the test prompt, which aggregates both training information from $\bv_s$ and test information from the test image $\x_t$ and class names $\mathcal{Y}_t$. The test prompt exploits the relationships between training and test distributions by the transformer inference network $\btheta$.
$\bv_t$ is then utilized into the frozen image and text encoders $\Phi = \{ \Phi_I, \Phi_T \} $ to generalize the CLIP model to the test data.

To optimize the model for generating the probabilistic training and test prompts, we further introduce variational inference to approximate the true posterior $p(\bv_t, \bv_s|\mathcal{D}_t, \mathcal{Y}_t, \mathcal{D}_s)$ into eq.~(\ref{eq: prior_predict}), which is factorized as:
\begin{equation}
q_{\btheta}(\bv_t, \bv_s|\mathcal{D}_t,\mathcal{Y}_t, \mathcal{D}_s) {=} q_{\btheta}(\bv_t|\bv_s, \mathcal{D}_t, \mathcal{Y}_t) p(\bv_s|\mathcal{D}_s),
\label{eq: varationalp}
\end{equation}
where $\mathcal{D}_t$ consists of test input-output pairs sampled from the test distribution $p(\x_t, \y_t)$. The variational posterior shares the same inference model $\btheta$ with the prior distribution.
By integrating eq.~(\ref{eq: varationalp}) into eq.~(\ref{eq: prior_predict}), the evidence lower bound (ELBO) of the log-likelihood $\log p_{\Phi, \btheta}(\y_{t}|\x_{t}, \mathcal{Y}_{t}, \mathcal{D}_s)$ is derived as: 
\begin{equation}
\small
\label{eq: elbo_app}
\begin{aligned}
    & \log p_{\Phi, \btheta}(\y_{t}|\x_{t}, \mathcal{Y}_{t}, \mathcal{D}_s) \\
    & = \log \int \int p(\y_t|\x_t, \bv_t, \mathcal{Y}_t) p(\bv_t, \bv_s|\x_t, \mathcal{Y}_t, \mathcal{D}_s) d\bv_t d\bv_s \\
    & = \log \int \int p(\y_{t'}|\x_{t}, \bv_{t}, \mathcal{Y}_{t}) q_{\btheta}(\bv_{t}, \bv_s|\mathcal{D}_t,\mathcal{Y}_t, \mathcal{D}_s) \\
    & ~~~~~~~~~~~ \frac{p(\bv_{t}, \bv_s|\x_t, \mathcal{Y}_t, \mathcal{D}_s)}{q(\bv_{t}, \bv_s|\mathcal{D}_t,\mathcal{Y}_t, \mathcal{D}_s)}  d\bv_t d\bv_s \\
    & = \log \int \int p(\y_{t'}|\x_{t}, \bv_{t}, \mathcal{Y}_{t}) q_{\btheta}(\bv_{t}, \bv_s|\mathcal{D}_t,\mathcal{Y}_t, \mathcal{D}_s) \\
    & ~~~~~~~~~~~ \frac{p_{\btheta}(\bv_t|\bv_s, \x_t, \mathcal{Y}_t) p(\bv_s|\mathcal{D}_s)}{q_{\btheta}(\bv_t|\bv_s, \mathcal{D}_t, \mathcal{Y}_t) p(\bv_s|\mathcal{D}_s)}  d\bv_t d\bv_s \\
    & = \log \int \int p(\y_{t'}|\x_{t}, \bv_{t}, \mathcal{Y}_{t}) q_{\btheta}(\bv_{t}, \bv_s|\mathcal{D}_t,\mathcal{Y}_t, \mathcal{D}_s) \\
    & ~~~~~~~~~~~ \frac{p_{\btheta}(\bv_t|\bv_s, \x_t, \mathcal{Y}_t)}{q_{\btheta}(\bv_t|\bv_s, \mathcal{D}_t, \mathcal{Y}_t)}  d\bv_t d\bv_s \\
    & 
    \geq \mathbb{E}_{q_{\btheta}(\bv_{t}, \bv_s)} \big [ \log p_{\Phi}(\y_t | \x_t, \bv_{t}, \mathcal{Y}_{t}) \big ] \\
    & ~~~~~~~~~~~ - \mathbb{D}_\mathrm{KL} \big [ q_{\btheta}(\bv_{t}|\bv_s, \mathcal{D}_t, \mathcal{Y}_{t}) || p_{\btheta}(\bv_{t}|\bv_s, \x_{t}, \mathcal{Y}_{t})  \big ],
\end{aligned}
\end{equation}
where the expectation of the log-likelihood is calculated on the variational posterior distribution $q_{\btheta}(\bv_t, \bv_s|\mathcal{D}_t, \mathcal{Y}_t, \mathcal{D}_s)$. 

Our goal is to maximize the log-likelihood of the test data $\log p_{\Phi, \btheta}(\y_{t}|\x_{t}, \mathcal{Y}_{t}, \mathcal{D}_s)$, i.e., maximize the ELBO in eq.~(\ref{eq: elbo_app}), which is equivalent to minimize the negative log-likelihood. 
Therefore, minimizing the loss function to optimize our any-shift prompting becomes minimizing:
\begin{equation}
\small
\label{eq: finalloss}
\begin{aligned}
    & - \log p_{\Phi, \btheta}(\y_{t}|\x_{t}, \mathcal{Y}_{t}, \mathcal{D}_s) \\
    & ~~~~~~~~~~~
    \leq \mathbb{E}_{q_{\btheta}(\bv_{t}, \bv_s)} \big [ - \log p_{\Phi}(\y_t | \x_t, \bv_{t}, \mathcal{Y}_{t}) \big ] \\
    & ~~~~~~~~~~~ + \mathbb{D}_\mathrm{KL} \big [ q_{\btheta}(\bv_{t}|\bv_s, \mathcal{D}_t, \mathcal{Y}_{t}) || p_{\btheta}(\bv_{t}|\bv_s, \x_{t}, \mathcal{Y}_{t})  \big ].
\end{aligned}
\end{equation}

\section{Details of setting and implementations}

\subsection{Details of datasets and settings}
\textbf{Covariate shift.}
We conduct the experiments on covariate shifts in two settings, multiple training distributions and single training distributions. 
The experiments on multiple training distributions are conducted on domain generalization datasets \texttt{PACS}, \texttt{VLCS}, \texttt{Office-Home}, and \texttt{DomainNet}, which contain multiple domains of images with the same label space. 
\texttt{PACS} \cite{li2017deeper} includes images of 7 classes from four different domains,  \textit{photo}, \textit{art-painting}, \textit{cartoon}, and \textit{sketch}.
\texttt{VLCS} \cite{fang2013video} consists of images of 5 classes and four different datasets, \textit{Pascal-VOC2007} \cite{everingham2010pascal}, \textit{LabelMe} \cite{russell2008labelme}, \textit{Caltech101} \cite{griffin2007caltech}, and \textit{SUN} \cite{choi2010exploiting}.  
\texttt{Office-Home} also contains four domains, \textit{art}, \textit{clipart}, \textit{product}, and \textit{real-world}, while the images are from 65 categories, which is much more than \texttt{PACS} and \texttt{VLCS}.
\texttt{DomainNet} is even larger, which consists of images from six domains and 345 categories. The domains are \textit{clipart}, \textit{inforgraph}, \textit{painting}, \textit{quickdraw}, \textit{real}, and \textit{sketch}.
We follow the ``leave-one-out protocol'' \cite{li2017deeper} on these datasets, where we select one domain as the test distribution, and the other domains are treated as the training distributions.
The model is trained on the training distributions and evaluated on the test one.
We treat each domain at the test distribution individually for evaluation and report the averaged results on all test distributions in Table 2 in the main paper. The detailed results of each test distribution are reported in the following section.

The experiments on single training distribution follow the domain generalization in Zhou \etal \cite{zhou2022conditional}, where the model is trained on \texttt{ImageNet} (1,000 categories) and evaluated on the other four variants \texttt{ImageNet-V2} \cite{recht2019imagenet}, \texttt{ImageNet-(S)ketch} \cite{wang2019learning}, \texttt{ImageNet-A} \cite{hendrycks2021natural}, and \texttt{ImageNet-R} \cite{hendrycks2021many} with the same label space.
Most of the above datasets have shifts in the images, i.e., marginal input distributions $p(\x)$. Therefore, we use these datasets for the evaluation of our method across covariate shift. 

\vspace{2mm}
\noindent
\textbf{Label shift.}
We conduct the experiments on label shift following the base-to-new classification setting in Zhou \etal \cite{zhou2022learning}.
In this case, the distribution shifts occur in the marginal output distribution $p(\y)$, where the ``new'' classes have $p(\y_c){=}0$ during training. 
We use eleven benchmarks with label shift. The benchmarks includes general classification datasets \texttt{ImageNet} \cite{deng2009imagenet} and \texttt{Caltech101} \cite{fei2004learning}; fine-grained classification datasets \texttt{OxfordPets} \cite{parkhi2012cats}, \texttt{StanfordCars} \cite{krause20133d}, \texttt{Flowers102} \cite{nilsback2008automated}, \texttt{Food101} \cite{bossard2014food}, and \texttt{FGVCAircraft} \cite{maji2013fine}; scene recognition dataset \texttt{SUN397} \cite{xiao2010sun}; action recognition dataset \texttt{UCF101} \cite{soomro2012ucf101}; texture classification dataset \texttt{DTD} \cite{cimpoi2014describing}; and satellite image recognition \texttt{EuroSAT} \cite{helber2019eurosat}.
We follow the same base-new classes split and evaluation set in Zhou \etal \cite{zhou2022conditional}.

\vspace{2mm}
\noindent
\textbf{Concept shift.}
We approximate the concept shift by relabeling the \texttt{ImageNet} dataset with the superclasses in \cite{santurkar2020breeds}. The model is trained on the original classes and evaluated on the superclasses. In this case, the marginal input distribution $p(\x)$ is the same while the conditional distributions $p(\y|\x)$ are different between training and test data. 

\vspace{2mm}
\noindent
\textbf{Conditional shift.} 
For conditional shift, we evaluate the proposed method on two subpopulation datasets, \texttt{Living-17} and \texttt{Entity-30} \cite{santurkar2020breeds}, which contain images of 17 animal categories and images of 30 entities, respectively. 
We follow the training and test split in \cite{garg2023rlsbench}, where the training and test distributions have the same overall classes but contain different subpopulations of those classes.
In this case, the marginal output distributions $p(\y)$ of training and test data are the same, while the input distributions are changed according to different categories, i.e., $p(\x|\y)$ are different. Therefore, we treat the setting as conditional shift.

\vspace{2mm}
\noindent
\textbf{Joint shift.}
To evaluate the proposed method on joint shift, we conduct experiments on \texttt{Office-Home} under the open domain generalization setting \cite{shu2021open}, which we refer to as \texttt{Open-Office-Home}. We split the label space of the 65 classes and make various label spaces across different domains. The split of classes is shown in Table~\ref{tab:split}.
Therefore, there are both covariate shift and label shift between the training and test distributions, which we treat as the joint shift on $p(\x, \y)$.

\begin{table}[t]
\centering
	\resizebox{0.9\columnwidth}{!}{%
\begin{tabular}{lc}
\toprule
Domains &  Classes \\ \midrule
Source 1 & 0 \-- 2, 3 \-- 8, 9 \-- 14, 21 \-- 31 \\
Source 2 & 0 \-- 2, 3 \-- 8, 15 \-- 20, 32 \-- 42 \\
Source 3 & 0 \-- 2, 9 \-- 14, 15 \-- 20, 43 \-- 53\\
Target & 0 \-- 64 \\ 
\bottomrule
\end{tabular}
}
\vspace{-2mm}
\caption{
\textbf{Classes split for joint distribution shifts} on \texttt{Open-Office-Home}. We use the numbers to denote the class names. The setting contains both covariate and label shifts, leading to joint shifts on $p(\x, \y)$.
}\label{tab:split}
\vspace{-4mm}
\end{table}


\subsection{Implementations and hyperparameters}

\begin{figure}[t]
\centering
\includegraphics[width=0.99\linewidth]{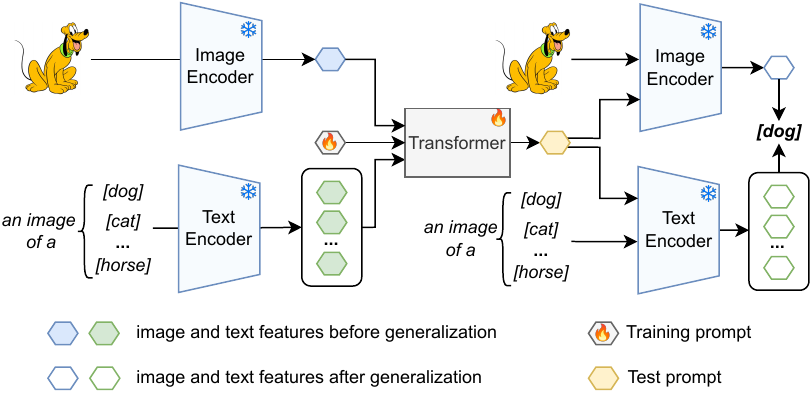}
\vspace{-2mm}
\caption{\textbf{Overall framework of generating the any-shift prompt and generalizing the CLIP model.}}
\label{fig: test}
\vspace{-3mm}
\end{figure}

For all experiments, we train and evaluate the model on a single NVIDIA V100 GPU. We use the same backbone and transformer inference network for all datasets. The backbone is the frozen CLIP model with ViT-B/16 as the image encoder. The transformer inference network consists of a 2-layer transformer and 2 MLP layers to generate the distribution of the test prompt. There are also two trainable vectors as the mean and variance of the probabilistic training prompt and trainable position embeddings for image and text features respectively. 
The sampled test prompt is then fed into both the image and text encoders to generalize the features and classifiers. We provide an illustration in Figure \ref{fig: test}.
Note that the test prompt is utilized as tokens of the image and text encoders. To make it the same size as the inputs, we use two linear layers to project the test prompt to the image path and text embedding space, respectively.

\begin{table}[t]
\centering
%
\resizebox{\linewidth}{!}{%
\begin{tabular}{l | ccccccccccc}
\toprule 
& \texttt{\rot{ImageNet}} & \texttt{\rot{Caltech101}} & \texttt{\rot{OxfordPets}} & \texttt{\rot{StanfordCars}} & \texttt{\rot{Flowers102}} & \texttt{\rot{Food101}} & \texttt{\rot{FGVC}} & \texttt{\rot{SUN397}} & \texttt{\rot{DTD}} & \texttt{\rot{EuroSAT}} & \texttt{\rot{UCF101}} \\ \midrule
Learning rate    & \multicolumn{11}{c}{$2e-3$} \\
Optimizer    & \multicolumn{11}{c}{SGD} \\ \midrule
Batch Size              & 1 & 4 & 8 & 6 & 4 & 4 & 4 & 2 & 8 & 10 & 4  \\
Epochs            & 10 & 30 & 30 & 30 & 30 & 30 & 30 & 30 & 30 & 30 & 30 \\
\bottomrule
\end{tabular}
}
\caption{\textbf{Dataset-specific hyper-parameters for label shift datasets and ImageNet-based datasets}. The ImageNet-based covariate shift, label shift, and concept shift datasets use the same hyperparameters.}
\vspace{-4mm}
\label{tab:dataset_hparams}
\end{table}

\begin{table}[!h]
\centering
%
\resizebox{\linewidth}{!}{%
\begin{tabular}{l | cccc c | cc}
\toprule 
& \texttt{\rot{PACS}} & \texttt{\rot{VLCS}} & \texttt{\rot{Office-Home}} & \texttt{\rot{Open-Office-Home}} & \texttt{\rot{DomainNet}} & \texttt{\rot{Living-17}} & \texttt{\rot{Entity-30}} \\ \midrule
Learning rate    & \multicolumn{7}{c}{$5e-4$} \\
Optimizer    & \multicolumn{7}{c}{Adam} \\ \midrule
Training iterations     & \multicolumn{4}{c}{3,000 iterations} & 10,000 iterations & \multicolumn{2}{c}{30 epochs}  \\ 
Batch Size              & 32 & 32 & 8 & 8 & 2 & 32 & 16 \\
\bottomrule
\end{tabular}
}
\caption{\textbf{Dataset-specific batch sizes for common domain generalization datasets and conditional shift datasets.}
}
\vspace{-2mm}
\label{tab:dataset_hparams2}
\end{table}

\begin{table}[t]
\begin{center}
	\resizebox{1\columnwidth}{!}{%
\begin{tabular}{lllllll}
\toprule
 &  & \multicolumn{5}{c}{Accuracy} \\
\cmidrule(lr){3-7}
Method & Iterations & Art & Clipart & Product & Real &\textit{\textbf{Mean}} \\ \midrule
CLIP baseline & - & 79.32 & 67.70 & 86.93 & 87.46 & 80.35\\ \midrule
Transformer adapter & 20,000 & 78.76 & 64.62 & 87.98 & 84.83 & 79.05 \\
\rowcolor{lightblue}
\textit{\textbf{Any-shift prompt}} & 3,000 & \textbf{83.40} & \textbf{72.53}  & \textbf{91.24} &  \textbf{90.84} & \textbf{84.50} \\ 
\bottomrule
\end{tabular}
}
\end{center}
\vspace{-5mm}
\caption{\textbf{Benefits of generalization with any-shift prompting.}
Directly training a transformer as an adapter of the image and textual features still easy to lead to overfitting. By aggregating the training, test, and relationship information into the prompt, any-shift prompting achieves better generalization.}
\label{transadapter}
\vspace{-3mm}
\end{table}

\begin{table}[t]
\begin{center}
	\resizebox{1\columnwidth}{!}{%
\begin{tabular}{llllll}
\toprule
Inference network   & Art & Clipart & Product & Real &\textit{\textbf{Mean}} \\ \midrule
CLIP baseline & 79.32 & 67.70 & 86.93 & 87.46 & 80.35\\ \midrule
Averaging & 82.27 & 70.91 & 89.95 & 89.66 & 83.20 \\
MLP & 82.48 & 71.09 & 90.18 & 89.73 & 83.37 \\
\rowcolor{lightblue}
\textit{\textbf{Transformer}}  & \textbf{83.40} & \textbf{72.53}  & \textbf{91.24} &  \textbf{90.84} & \textbf{84.50} \\ 
\bottomrule
\end{tabular}
}
\end{center}
\vspace{-5mm}
\caption{\textbf{Ablations on the aggregation methods.}
The transformer inference network performs best since it better encodes the relationships between different information.}
\label{ablationt}
\vspace{-4mm}
\end{table}

\begin{table*}[t]
    \resizebox{2.09\columnwidth}{!}{%
		\setlength\tabcolsep{4pt} 
    \begin{tabular}{l c ccccccccccc}
    \toprule
    & \textbf{Source} & \multicolumn{11}{c}{\textbf{Target}} \\ \cmidrule(lr){2-2} \cmidrule(lr){3-13}
    & ImageNet & Caltech101 & OxfordPets & StanfordCars & Flowers102 & Food101 & FGVCAircraft & SUN397 & DTD & EuroSAT & UCF101 & \emph{Average} \\
    \midrule
    CoOp~\cite{zhou2022learning} & \textbf{71.51} & 93.70 & 89.14 & 64.51 & 68.71 & 85.30 & 18.47 & 64.15 & 41.92 & {46.39} & 66.55 & 63.88 \\
    CoCoOp~\cite{zhou2022conditional} & 71.02 & 94.43 & {90.14} & {65.32} & {71.88} & {86.06} & {22.94} & {67.36} & {45.73} & {45.37} & {68.21} & {65.74} \\
    TPT~\cite{shu2022test} & 68.98 & 68.98 & 47.75 & 87.79 & 66.87 & 68.04 & 94.16 & 84.67 & 65.50 & 24.78 & 42.44 & 65.10 \\
    BPL~\cite{derakhshani2023bayesian} & 70.70 & 93.67 & 90.63 & 65.00 & 70.90 & \textbf{86.30} & 24.93 & \textbf{67.47} & 46.10 & 45.87 & 68.67 & 65.95 \\
    MaPLe~\cite{khattak2023maple} & 70.72 & 93.53 & 90.49 & 65.57 & 72.23 & 86.20 & 24.74 & 67.01 & 46.49 & 48.06 & 68.69 & 66.30 \\
    \rowcolor{lightblue}
    \textit{\textbf{This paper}} & 71.05 & \textbf{94.57} & \textbf{90.79} & \textbf{66.90} & \textbf{72.30} & 86.17 & \textbf{25.16} & 67.32 & \textbf{47.35} & \textbf{50.25} & \textbf{69.52} & \textbf{67.03} \\ 
    \bottomrule
    \end{tabular}}
    \caption{\textbf{Comparison of prompt learning methods in the cross-dataset transfer setting.} Our method achieves the best overall performance on 10 test datasets.
    }
    \vspace{-5mm}
    \label{tab:xd}
\end{table*}

Except for the architecture and settings shared by all datasets, we also provide the specific hyperparameters for different datasets.
Batch size is a hyperparameter that varies per dataset (Tables \ref{tab:dataset_hparams} and \ref{tab:dataset_hparams2}).
For the experiments of label shift (eleven datasets) and the others based on \texttt{ImageNet} (ImageNet-based covariate shift and concept shift), we use the same learning rate $2e-3$ as Zhou \etal \cite{zhou2022conditional} with SGD. The dataset-specific batch size and epochs are provided in Table \ref{tab:dataset_hparams}.
For the covariate shift datasets \texttt{PACS}, \texttt{VLCS}, \texttt{Office-Home}, \texttt{DomainNet} and joint shift dataset \texttt{Open-Office-Home}, we train the model with $5e-4$ learning rate and 3000 iterations by Adam optimizer. 
For the conditional shift dataset conditional shift datasets \texttt{Living-17} and \texttt{Entity-30}, we use the same learning rate $5e-4$ and Adam optimizers for 30 epochs. The details are shown in Table \ref{tab:dataset_hparams2}.

\section{More ablations and comparisons}

\paragraph{Benefits of generalization with prompts}
In our any-shift prompting, we generate the test prompt by aggregating the training information and the test information by a transformer inference network. The test information is from the image and textual features of the CLIP model. 
In addition to generating the prompt for the CLIP model, another way to achieve generalization is directly adapting the image and textual features by the transformer network and making predictions by the image and textual features. 
To show the benefits of generalization with our any-shift prompting, we conduct an experiment that adapts the image and textual features using the same transformer inference network, which we refer to as ``Transformer adapter''.
The experimental results on \texttt{Open-Office-Home} are reported in Table \ref{transadapter}. The transformer adapter performs even worse than the CLIP baseline since it is still easy to overfit the training distribution.
Moreover, the transformer adapter requires much more training costs (20,000 iterations) than any-shift prompting (3,000 iterations).
The results demonstrate both the effectiveness and efficiency of our any-shift prompting for generalization across distribution shifts.

\begin{table}[t]
\begin{center}
	\resizebox{1\columnwidth}{!}{%
\begin{tabular}{llllll}
\toprule
 Method   & Photo & Art & Cartoon & Sketch &\textit{\textbf{Mean}} \\ \midrule
CLIP & 99.94 & 97.41 & 98.98 & 88.19 & 96.13\\
CLIP-D & 99.94 & 97.61 & 99.02 & 90.03 & 96.65 \\
CoOp & 99.70 & 97.56 & 98.59 & 89.95 & 96.45 \\
CoCoOp & 99.94 & 98.09 & 99.19  & 90.77 & 97.00 \\
TPT & 99.82 & 97.68 & 98.92  & 92.58 & 97.25 \\
\rowcolor{lightblue}
\textit{\textbf{This paper}}  & \textbf{99.94} & \textbf{98.86} & \textbf{99.32} &  \textbf{94.53} & \textbf{98.16} \scriptsize{$\pm$ 0.4}\\
\bottomrule
\end{tabular}
}
\end{center}
\vspace{-5mm}
\caption{\textbf{Detailed comparisons on \texttt{PACS} with covarate shift.}}
\label{pacs}
\end{table}

\begin{table}[t]
\begin{center}
	\resizebox{1\columnwidth}{!}{%
\begin{tabular}{llllll}
\toprule
 Method   & VOC & LabelMe & Caltech & SUN &\textit{\textbf{Mean}} \\ \midrule
CLIP & 84.32 & 68.26 & 98.61 & 74.52 & 81.43\\
CLIP-D & 82.60 & 68.76 & 98.76 & 72.68 & 80.70 \\
CoOp & 85.86 & 68.51 & 98.94 & 76.72 & 82.51 \\
CoCoOp & 86.03 & 70.45 & 99.12 & 77.96 & 83.39 \\
TPT & 86.20 & 71.05 & 99.46 & 80.60 & 84.33 \\
\rowcolor{lightblue}
\textit{\textbf{This paper}}  & \textbf{88.14} & \textbf{72.65} & \textbf{100.00} &  \textbf{85.37} & \textbf{86.54} \scriptsize{$\pm$ 0.4}\\ 
\bottomrule
\end{tabular}
}
\end{center}
\vspace{-5mm}
\caption{\textbf{Detailed comparisons on \texttt{VLCS} with covarate shift.}}
\label{vlcs}
\end{table}

\begin{table}[t]
\begin{center}
	\resizebox{1\columnwidth}{!}{%
\begin{tabular}{llllll}
\toprule
 Method   & Art & Clipart & Product & Real &\textit{\textbf{Mean}} \\ \midrule
CLIP & 79.32 & 67.70 & 86.93 & 87.46 & 80.35\\
CLIP-D & 80.47 & 68.83 & 87.93 & 88.80 & 81.51 \\
CoOp & 80.99 & 69.52 & 88.69 & 89.28 & 82.12 \\
CoCoOp & 81.78 & 70.09 & 89.32  & 89.89 & 82.77 \\
TPT & 82.45 & 71.18 & 90.03  & 90.15 & 83.45 \\
\rowcolor{lightblue}
\textit{\textbf{This paper}}  & \textbf{83.70} & \textbf{73.00} & \textbf{92.50} &  \textbf{91.44} & \textbf{85.16} \scriptsize{$\pm$ 0.6}\\ 
\bottomrule
\end{tabular}
}
\end{center}
\vspace{-5mm}
\caption{\textbf{Detailed comparisons on \texttt{Office-Home}.}}
\label{office}
\end{table}

\begin{table}[t]
\begin{center}
	\resizebox{1\columnwidth}{!}{%
\begin{tabular}{llllllll}
\toprule
 Method    & Clipart & Painting & Real & Infograph & Quickdraw & Sketch &\textit{\textbf{Mean}} \\ \midrule
CLIP & 68.12 & 56.18 & 78.82  & 46.36 & 14.32 & 60.69 & 54.08 \\
CLIP-D & 70.83 & 58.02 & 80.52 & 48.85 & 16.39 & 62.84 & 56.24 \\
CoOp & 74.39 & 61.18 & 83.26 & 51.88 & 16.67 & 65.52 & 58.82 \\
CoCoOp & 74.82 & 61.56 & 83.98  & \textbf{52.68} & 17.47 & 66.10 & 59.43 \\
TPT & 75.09 & 62.77 & 84.67  & 52.65 & 17.28 & 66.98 & 59.90 \\
\rowcolor{lightblue}
\textit{\textbf{This paper}}  & \textbf{76.08} & \textbf{66.62} & \textbf{85.03} &  52.56 & \textbf{18.05} & \textbf{67.26} & \textbf{60.93} \scriptsize{$\pm$ 0.4} \\ 
\bottomrule
\end{tabular}
}
\end{center}
\vspace{-5mm}
\caption{\textbf{Detailed comparisons on \texttt{DomainNet}.}}
\label{domnet}
\end{table}

\vspace{2mm}
\noindent
\textbf{Benefits of the transformer inference network}
We also conduct experiments on \texttt{Open-Office-Home} with different methods for aggregating the training and test information.
We generate the test prompt by directly averaging the training prompts, the test image feature, and textual features. 
In addition, we also use an MLP network to replace the transformer network to generate the test prompt from the averaged features. 
As shown in Table \ref{ablationt}, the transformer inference network achieves the best performance, demonstrating the effectiveness of considering the relationships between different information for aggregation.


\vspace{2mm}
\noindent
\textbf{Comparison on cross-dataset shift.}
Following Zhou \etal \cite{zhou2022conditional}, we conduct experiments on the cross-dataset setting, where the model trained on \texttt{ImageNet} is evaluated on the other 10 datasets shown in Table~\ref{tab:xd}. In this case, there are different distribution shifts for different test datasets. 
Compared with the other prompt learning methods, e.g., CoOp \cite{zhou2022learning}, CoCoOp \cite{zhou2022conditional}, BPL \cite{derakhshani2023bayesian}, MaPLe \cite{khattak2023maple}, and test-time tuning method TPT \cite{shu2022test}, our method shows improvement on 8 of the 10 datasets, as well as the averaged result.

\vspace{2mm}
\noindent
\textbf{Detailed results on covariate shift}
We also report the detailed comparisons of each test distribution on the four covariate shift datasets. The results of \texttt{PACS}, \texttt{VLCS}, \texttt{Office-Home}, and \texttt{DomainNet} are provided in Table \ref{pacs}, \ref{vlcs}, \ref{office}, and \ref{domnet}, respectively.
Our method achieves the best performance on most of the test distributions.

\vspace{2mm}
\noindent
\textbf{Inference efficiency.}
Since our method only uses a single feedforward pass for generating the test prompts and making predictions, the inference time cost per iteration on a single V100 GPU (0.13s) is slightly higher than other prompt tuning methods like CoOp (0.10s) and CoCoOp (0.11s), and faster than TPT (0.25s), which has 1-step optimization at test time. 


\end{document}